%% file: ms.tex
\documentclass[journal, 10pt]{IEEEtran}

\input{mypackages}

\input{glossaries.tex}

\input{matrix_commands.tex}
\newcommand{\COMMENT}[1]{}
\newcommand{\REVISED}[1]{#1}
\newcommand{\REVSPACE}{}  

\begin{document}
\title{Relational Fusion Networks:\\Graph Convolutional Networks for Road Networks}
\author{Tobias Skovgaard Jepsen, Christian S. Jensen, and Thomas Dyhre Nielsen \\
        \small{Department of Computer Science, Aalborg University, Denmark}
\thanks{
  \footnotesize
  \emph{\textcopyright{} 2020 IEEE.  Personal use of this material is permitted.  Permission from IEEE must be obtained for all other uses, in any current or future media, including reprinting/republishing this material for advertising or promotional purposes, creating new collective works, for resale or redistribution to servers or lists, or reuse of any copyrighted component of this work in other works.}
}
}
\IEEEpubid{}

\maketitle

\input{abstract.tex}
\color{black}

\input{introduction.tex}

\input{preliminaries.tex}
\input{method.tex}

\input{experiments_setup.tex}
\input{experiments_results.tex}
\input{related_work.tex}
\input{epilogue.tex}
\input{acknowledgements.tex}

\printbibliography
\vspace{-1.2cm}
\begin{IEEEbiography}[{\includegraphics[width=1in, height=1.25in, trim={9.8cm 3.5cm 5cm 3cm}, clip, keepaspectratio]{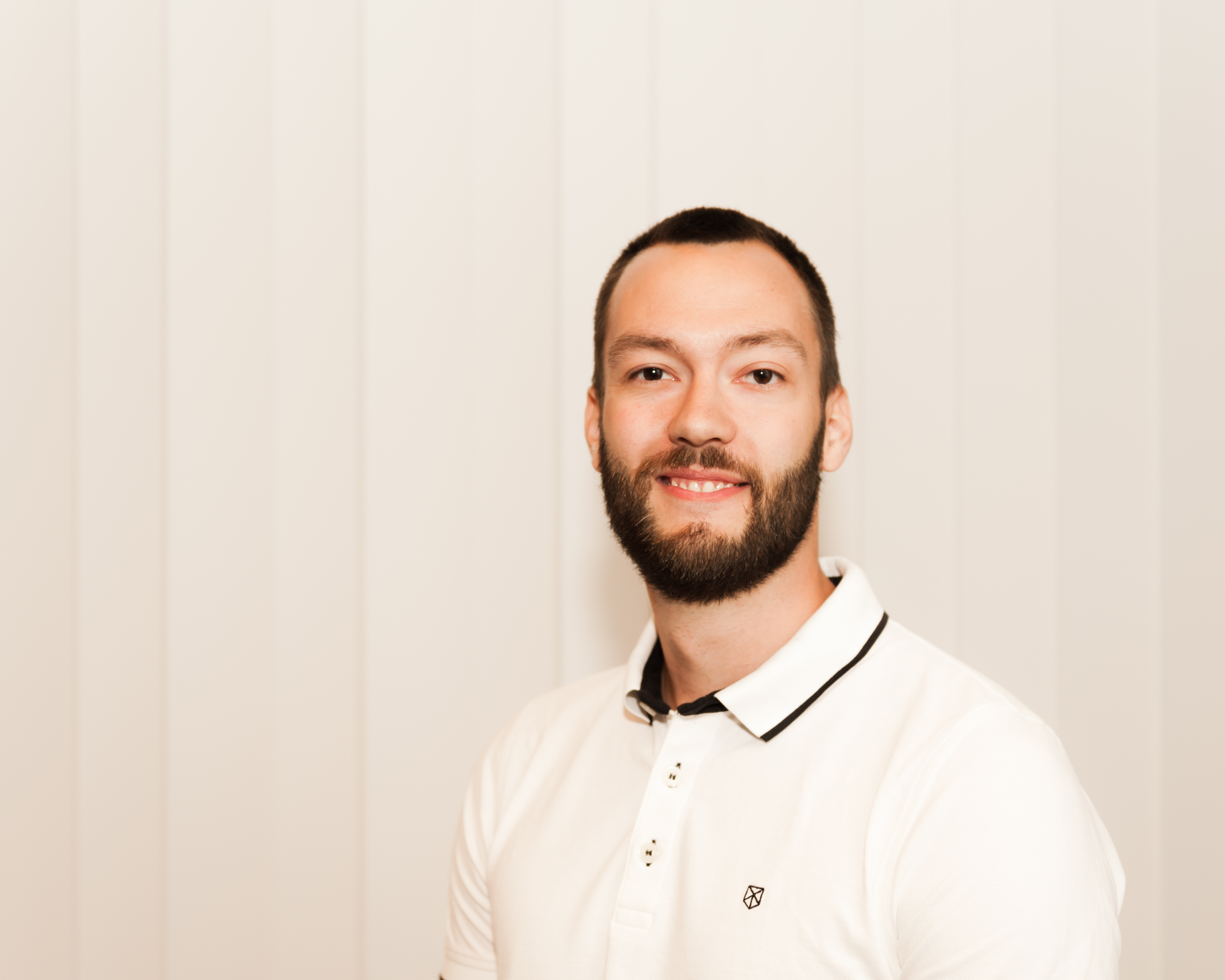}}]{Tobias Skovgaard Jepsen}
  is Ph.D. Fellow in Computer Science at Aalborg University, Denmark.
  He received his B.S.\ degree in software engineering in 2015 and his M.S.\ degree in computer science in 2017 from Aalborg University.
  His research interests concern the design and application of machine learning methods for transportation tasks.
\end{IEEEbiography}
\vspace{-1.1cm}
\begin{IEEEbiography}[{\includegraphics[width=1in, height=1.25in, trim={1cm 0 1cm 0}, clip, keepaspectratio]{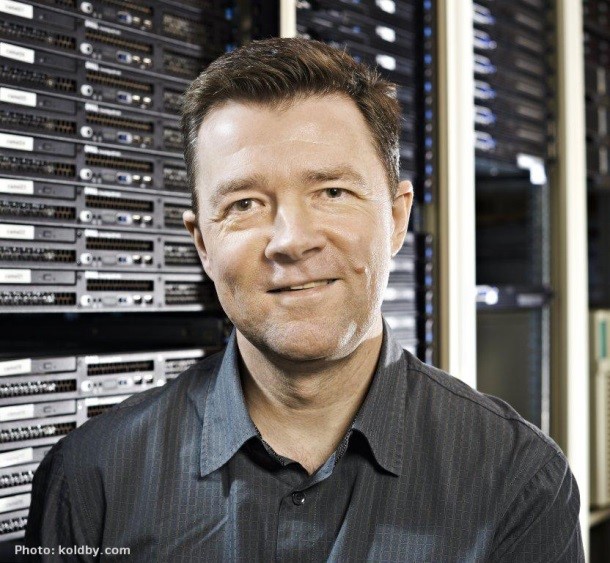}}]{Christian S. Jensen}
   is Professor of Computer Science at Aalborg University, Denmark.
   He was recently at Aarhus University for three years and at Google Inc.\ for one year.
   His research concerns data analytics and data-intensive systems, and its focus is on temporal and spatiotemporal analytics.
   Christian is an ACM and an IEEE fellow, and he is a member of the Academia Europaea, the Royal Danish Academy of Sciences and Letters, and the Danish Academy of Technical Sciences.
\end{IEEEbiography}
\vspace{-1.375cm}
\begin{IEEEbiography}[{\includegraphics[width=1in, height=1.25in, trim={1cm 0cm 1cm 0cm}, clip, keepaspectratio]{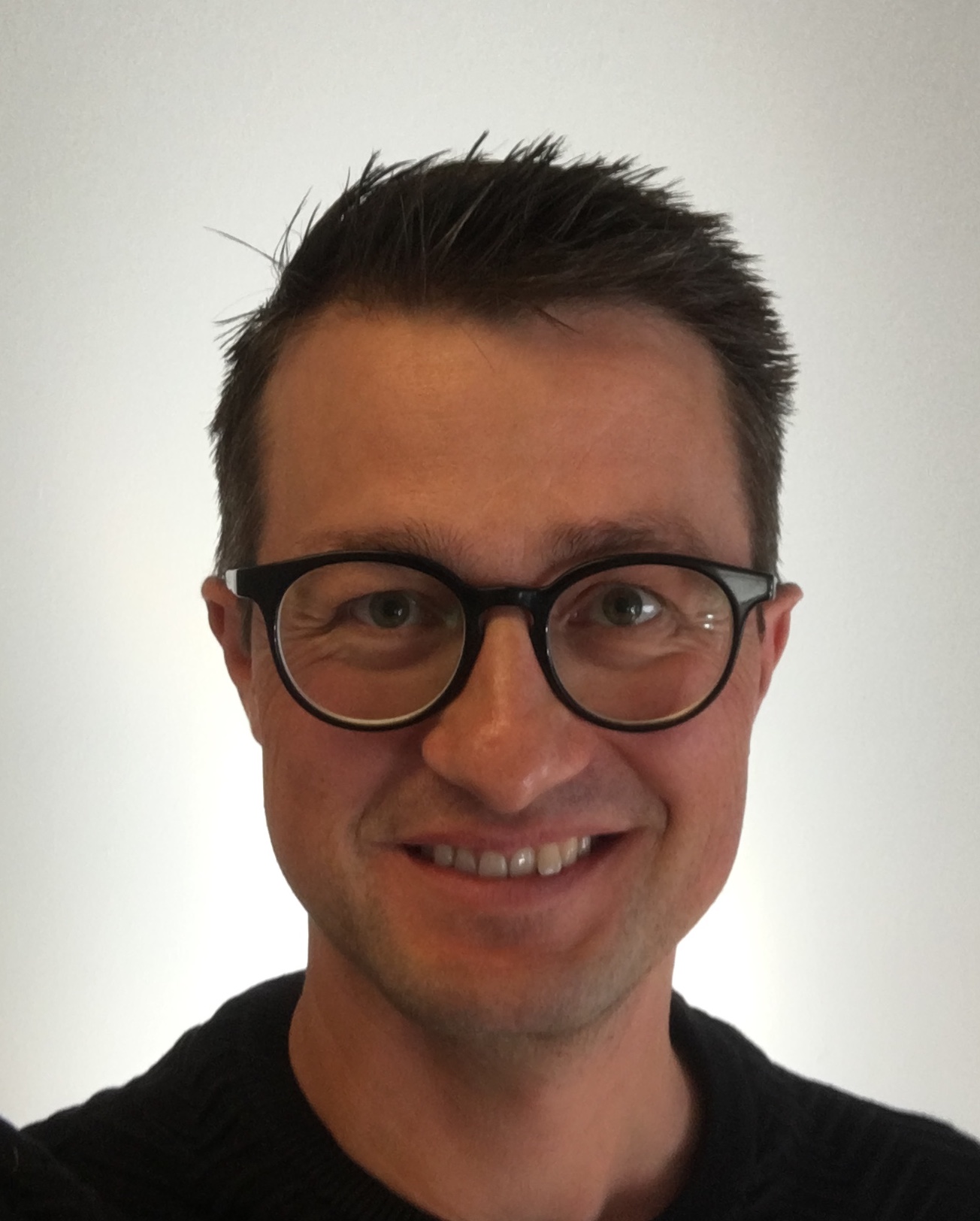}}]{Thomas Dyhre Nielsen}
  is Professor MSO (with special obligations) at Aalborg University, Denmark.
  His research interests concern different aspects of decision support systems, but with particular focus on machine learning and probabilistic graphical models including Bayesian networks.
  He is a co-author on a textbook on probabilistic graphical models and decision analysis and he is senior area editor for the International Journal of Approximate Reasoning.
\end{IEEEbiography}

\REVISED{
\begin{appendices}
\input{appendix.tex}
\end{appendices}
}
\end{document}

%% file: mypackages.tex
\usepackage[table]{xcolor}
\usepackage[]{todonotes}
\presetkeys{todonotes}{inline}{}
\usepackage{glossaries}
\usepackage{savesym}
\savesymbol{checkmark}
\usepackage{dingbat}
\usepackage{bbm}
\usepackage{array, makecell}
\usepackage{tikz}
\usepackage{pgfplots}

\makeatletter
\let\MYcaption\@makecaption
\makeatother

\usepackage[font=footnotesize]{subcaption}

\makeatletter
\let\@makecaption\MYcaption
\makeatother

\usepackage[utf8]{inputenc}
\usepackage[T1]{fontenc}

\usepackage{algorithm}
\usepackage{algpseudocode}
\usepackage{icomma}

\usepackage[backend=biber,style=ieee,natbib=true]{biblatex}
\usepackage{booktabs}
\usepackage[draft,bookmarks=false]{hyperref}

\usepackage{amsfonts}
\usepackage{textcomp}
\usepackage{calc}

\usepackage[capitalise]{cleveref}  
\creflabelformat{equation}{#2\textup{#1}#3}

\usepackage{multirow}

%% file: glossaries.tex
\usepackage{glossaries}

\newacronym{gps}
{GPS}
{Global Positioning System}

\newacronym{osm}
{OSM}
{OpenStreetMap}

\newacronym{rne}
{RNE}
{Road Network Embedding}

\newacronym{dfs}
{DFS}
{Depth-First Search}

\newacronym{elu}
{ELU}
{Exponential Linear Unit}

\newacronym{relu}
{ReLU}
{Rectified Linear Unit}

\newacronym{bfs}
{BFS}
{Breadth-First Search}

\newacronym{gcn}
{GCN}
{Graph Convolutional Network}

\newacronym{rfn}
{RFN}
{Relational Fusion Network}

\newacronym{mlp}
{MLP}
{Multi-Layer Perceptron}

\newacronym{tsne}
{t-SNE}
{t-distributed Stochastic Neighbor Embedding}

\newacronym{mape}
{MAPE}
{Mean Absolute Percentage Error}

\newacronym{mae}
{MAE}
{Mean Absolute Error}

\newacronym{mse}
{MSE}
{Mean Squared Error}

\newacronym{mpe}
{MPE}
{Mean Percentage Error}

%% file: abstract.tex
\begin{abstract}
  The application of machine learning techniques in the setting of road networks holds the potential to facilitate many important intelligent transportation applications.
  \glspl{gcn} are neural networks that are capable of leveraging the structure of a network.
  However, many implicit assumptions of \glspl{gcn} do not apply to road networks.
 
  We introduce the \emph{\glsfirst{rfn}}, a novel type of \gls{gcn} designed specifically for road networks.
  In particular, we propose methods that outperform state-of-the-art \glspl{gcn} by {$\mathbf{21}\boldsymbol{\%}$-$\mathbf{40}\boldsymbol{\%}$} on two machine learning tasks in road networks.
  Furthermore, we show that state-of-the-art \glspl{gcn} may fail to effectively leverage road network structure and may not generalize well to other road networks.
\end{abstract}

%% file: introduction.tex
\section{Introduction}\label{sec:introduction}
Machine learning on road networks can facilitate important intelligent transportation applications such as traffic flow prediction~\citep{traffic-flow-prediction, deep-embedded-traffic-flow-prediction}, traffic speed forecasting~\citep{spatio-temporal-gcn, long-term-traffic-speed-pred}, speed limit annotation~\citep{workshop}, and travel time estimation~\citep{multi-task-travel-time-estimation, bus-travel-time}.
However, machine learning on road networks is difficult due to the low number of attributes, often with missing values, that typically are available~\citep{workshop}.
This lack of attribute information can be alleviated by exploiting the network structure into the learning process~\citep{workshop}.
To this end, we propose the \emph{\glsfirst{rfn}}, a type of \glsfirst{gcn} designed for machine learning on road networks.

\glspl{gcn} are a type of neural network that operates on graph representations of networks.
\glspl{gcn} can in theory leverage the road network structure by aggregating over a road segment's neighborhood when computing the segment's representation, e.g., computing the mean representations of its adjacent road segments.

State-of-the-art \glspl{gcn} are designed to target node classification tasks in social, citation, and biological networks~\citep{bruna2014, chebnet, gcn, graphsage, gat, cayleynet}. Such networks differ substantially from road networks in terms of the attribute information available and other network characteristics. As a result, many implicit assumptions in \gls{gcn} proposals do not hold. 

First, road networks are \emph{edge-relational} and contain not only node and edge attributes, but also \emph{between-edge attributes} that characterize the relationship between road segments, i.e., the edges in a road network.
For instance, the angle between two road segments is informative for travel time estimation since it influences the time it takes to move from one segment to the other.

Second, compared to social, citation, and biological networks, road networks are low-density: road segments have few adjacent road segments.
For instance, the Danish road network has a mean node degree of $2.2$~\citep{workshop} compared to the mean node degrees $9.15$ and $492$ of a citation and a social networks, respectively~\citep{graphsage}.
The small neighborhoods in road networks make neighborhood aggregation in \glspl{gcn} sensitive to aberrant neighbors, which may contribute noise.

Third, \glspl{gcn} implicitly assume that the underlying network exhibits homophily meaning that adjacent road segments tend to be similar, and that changes in network characteristics, e.g., driving speeds, occur gradually.
Although road networks exhibit homophily, the homophily is volatile in the sense that regions can be highly homophilic, but have sharp boundaries characterized by abrupt changes in, e.g., driving speeds. This might for instance be caused by a change in speed limit such as when exiting a motorway.
In the most extreme case, a region may consist of a single road segment, \emph{in which case there is no homophily}.

\begin{figure}[t]
  \centering
    \includegraphics[trim={0cm 6cm 1.5cm 8cm},clip, width=0.95\columnwidth]{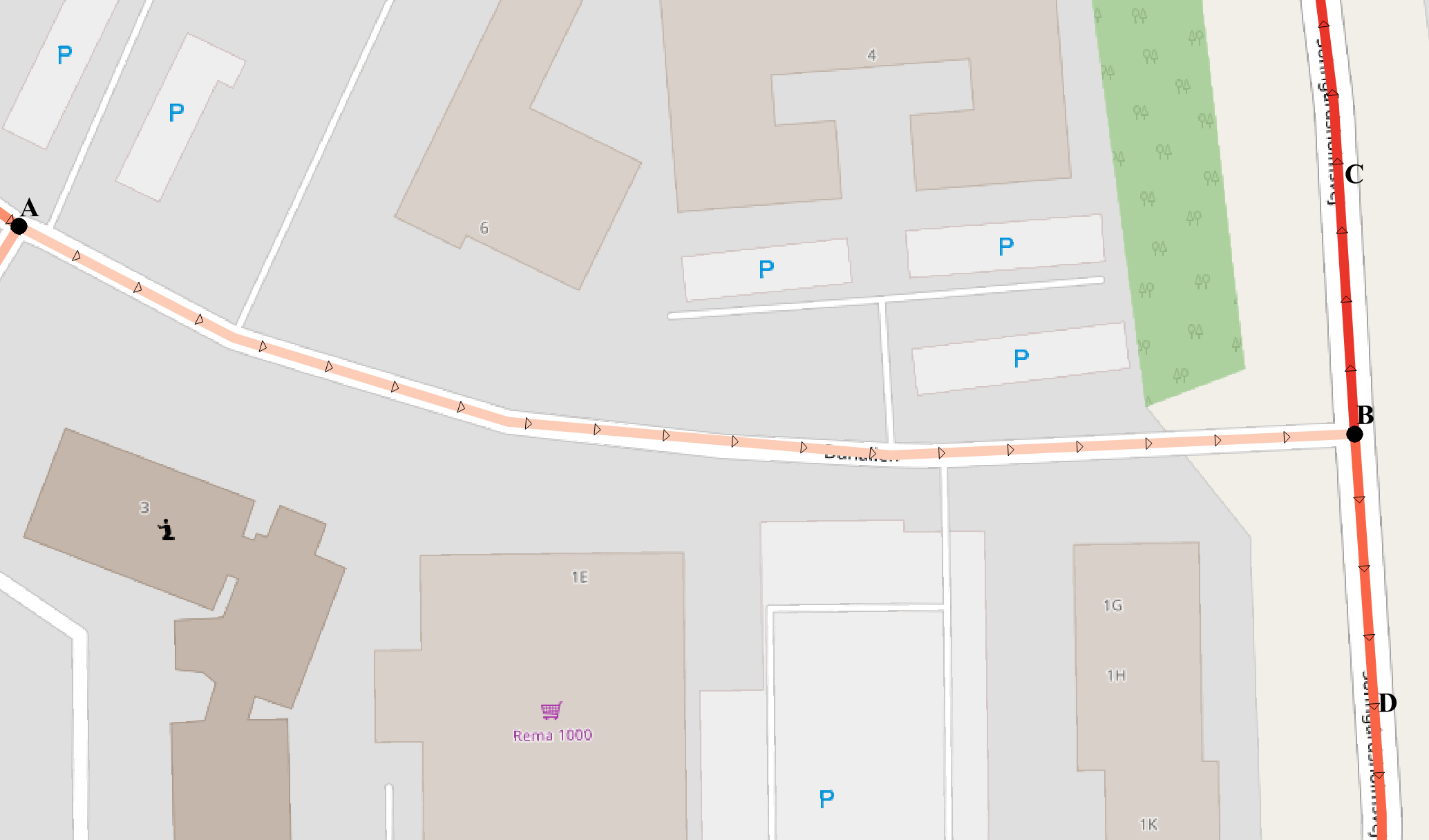}
  \caption{Two three-way intersections in Denmark. We illustrate the observed driving speeds for one driving direction per segment and color them accordingly: the darker, the faster the speed. Black dots mark intersections and triangles indicate driving directions.\label{fig:compelling-example}}
\end{figure}

As an example, the three-way intersection to the right in \cref{fig:compelling-example} exhibits of volatile homophily.
The two vertical road segments to the right in the figure and the road segments to the left in the figure form two internally homophilic regions: within each region, the road segments exhibit similar driving speeds.
The two regions are adjacent in the network, but a driver moving from one region to the other experiences an abrupt change in driving speed.

We suggest that new \gls{gcn} architectures are needed for high-performance machine learning on road networks based on our observations regarding current state-of-the-art \glspl{gcn}.
Addressing the challenge of volatile homophily is of particular importance.
We therefore propose the \glsfirst{rfn}, a novel \gls{gcn} architecture designed to be generally applicable to machine learning tasks on road networks.

\REVISED{
Unlike \glspl{gcn}, \glspl{rfn} take into account the inherent properties of road networks, i.e., that they are edge-relational, low-density, and exhibit volatile homophily.
Specifically, \glspl{rfn} (i) explicitly incorporate the relationships between edges during aggregation, (ii) use an attention mechanism to exclude noise-contributing neighbors during aggregation, and (iii) use a relational fusion operator that allows an \gls{rfn} to only conditionally rely on the homophily assumption when performing neighborhood aggregation.
}

\COMMENT{
\glspl{rfn} have two major advantages over state-of-the-art \gls{gcn}:
(i) \glspl{rfn} include a relational function operator that allows an \gls{rfn} to only conditionally rely on the homophily assumption when performing neighborhood aggregation, and 
(ii) \glspl{rfn} supports node, edge, and between-edge attributes as opposed to only node attributes.
In addition, we combine the above features with a recent advancement in \gls{gcn} architecture~\citep{gat} and design an attention mechanism for \glspl{rfn} that allow an \gls{rfn} to filter out any aberrant neighbors that would otherwise contribute noise to the neighborhood aggregate.
}

We experimentally evaluate different \gls{rfn} variants on two example machine learning tasks on road networks: driving speed estimation and speed limit classification.
Our experiments show that the best \gls{rfn} variants outperform state-of-the-art \glspl{gcn} by $32\text{--}40\%$ and $21\text{--}24\%$ on driving speed estimation and speed limit classification, respectively.

In our experiments, we demonstrate that state-of-the-art \glspl{gcn} fail to leverage road network structure on the speed limit classification task where \glspl{rfn} do not.
Finally, we show that the knowledge learned by an \gls{rfn} can generalize well to an entirely unseen part of the road network.
This feature is particular important in cases where there is a large spatial imbalance, e.g., when using crowd-sourced data or vehicle trajectory data.
In such data, data points tend to be concentrated in areas with high population density such as in major cities.

The remainder of the paper is structured as follows.
In \cref{sec:related-work}, we review related work.
In \cref{sec:preliminaries}, we give the necessary background on graph modeling of road networks and \glspl{gcn}.
In \cref{sec:method}, we describe \glspl{rfn} in detail.
In \cref{sec:experiments}, we describe our experiments and present our results.
Finally, we conclude in \cref{sec:conclusion}.

%% file: preliminaries.tex
\section{Preliminaries}\label{sec:preliminaries}
We now cover the necessary background in modeling road networks as graphs and \glspl{gcn}.

\subsection{Modeling Road Networks}\label{sec:road-network-modelling}
\begin{figure}
  \centering
  \begin{subfigure}{0.39\columnwidth}
    \centering
    \input{illustrations/primal_graph.tikz.tex}
    \vspace*{0cm}
    \caption{\scriptsize{Primal Graph.}\label{fig:graph-rep-primal}}
  \end{subfigure}
  \begin{subfigure}{0.59\columnwidth}
    \centering
    \vspace*{0.4805cm}
    \input{illustrations/dual_graph.tikz.tex}
    \vspace*{0.4805cm}
    \caption{\scriptsize{Dual Graph.}\label{fig:graph-rep-dual}}
  \end{subfigure}
  \caption{The (a) primal and (b) dual graph representations of the three-way intersection to the right in \cref{fig:compelling-example}.\label{fig:graph-rep}}
  \vspace*{-0.5cm}
\end{figure}
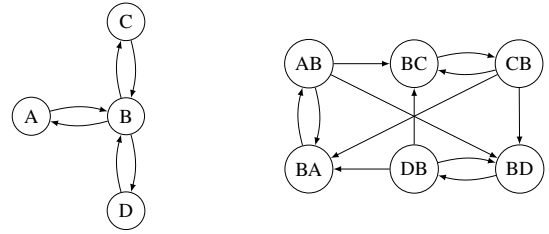

We model a road network as an attributed directed graph $G = (V, E, A^V, A^E, A^B)$ where $V$ is the set of nodes and $E$ is the set of edges.
Each node $v \in V$ represents an intersection, and each edge $(u, v) \in E$ represents a road segment that enables traversal from $u$ to $v$.
Next, $A^V$ and $A^E$ maps intersections and road segments, respectively, to their attributes. In addition, $A^B$ maps a pair of road segments $(u, v), (v, w) \in E$ to their between-segment attributes such as the angle between $(u, v)$ and $(v, w)$ based on their spatial representation.
An example of a graph representation of the three-way intersection to the right in figure \cref{fig:compelling-example} is shown in \cref{fig:graph-rep-primal}.

Two intersections $u$ and $v$ in $V$ are adjacent if there exists a road segment $(u, v) \in E$ or $(v, u) \in E $. 
Similarly, two road segments $(u_1, v_1)$ and $(u_2, v_2)$ in $E$ are adjacent if $v_1 = u_2$ or $v_2 = u_1$. 
The function $N\colon V \cup E \xrightarrow{} 2^V \cup 2^E$ returns the neighborhood, i.e., the set of all adjacent intersections or road segments, of a road network element $g \in V \cup E$.
The dual graph representation of $G$ is then $G^D=(E, B)$ where $B = \big\{ \big((u, v), (v, w) \big) \mid (u, v), (v, w) \in E \big\}$ is the set of \emph{between-edges}.
Thus, $E$ and $B$ \REVISED{serve as the node and edge sets of the dual graph, respectively}. 
For disambiguation, we refer to $G$ as the primal graph representation.

\subsection{Graph Convolutional Networks}\label{sec:gcn}
A \gls{gcn} is a neural network that operates on graphs and consists of one or more graph convolutional layers.
A graph convolutional network takes as input a graph $G = (V, E)$ and a numeric node feature matrix $\mathbf{X}^V \in \mathbb{R}^{|V| \times d_{\mathit{in}}}$, where each row corresponds to a $d_{\mathit{in}}$-dimensional vector representation of a node. 
Given these inputs, a \gls{gcn} computes an output at a layer $k$ s.t.\
\begin{equation}\label{eq:gcn-propagation}
  \mathbf{H}^{(V, k)}_v = \sigma(\textsc{Aggregate}^k(v) \mathbf{W}^k)\text{,}
\end{equation}
where $\sigma$ is an activation function,
$\textsc{Aggregate}\colon 2^V \rightarrow \mathbb{R}^{d_{\mathit{in}}}$ is a neighborhood aggregation function, and
$\mathbf{W}^k \in \mathbb{R}^{d_{\mathit{in}} \times d_o}$ is a learned weight matrix.
Similarly to $\mathbf{X}^V$, each row in $\mathbf{H}^{(V, k)}$ is a vector representation of a node.
Note that in some cases $\mathbf{X}^V$ is linearly transformed using matrix multiplication with a weight matrix $\mathbf{W}^k$ before aggregation~\citep{gat} while in other cases, weight multiplication is done after aggregation~\citep{gcn,graphsage}, as in \cref{eq:gcn-propagation}.

The \textsc{Aggregate} function in \cref{eq:gcn-propagation} derives a new representation of a node $v$ by aggregating over the representations of its neighbors.
While the aggregate function is what distinguishes \gls{gcn} architectures from each other, many can be expressed as a weighted sum~\citep{gcn,graphsage,gat}:
\begin{equation}\label{eq:aggregateWeightedSum}
  \textsc{Aggregate}^k(v) = \sum_{n \in N(v)} a_{(v,n)} \mathbf{H}^{(V, k-1)}_n\text{,}
\end{equation}
where $\mathbf{H}^{(V, 0)} = \mathbf{X}^V$, and $a_{(v,n)}$ is the aggregation weight for neighbor $n$ of node $v$.
For instance, $a_{(v, n)} = |N(v)|^{-1}$ in the mean aggregator of GraphSAGE~\citep{graphsage}.

%% file: illustrations/primal_graph.tikz.tex
\usetikzlibrary{shapes,arrows,positioning,fit,backgrounds}
\scalebox{0.75}{
\begin{tikzpicture}[align=center]
  \node [draw, circle] (A) {A};
  \node [draw, circle, right=1cm of A] (B) {B};
  \node [draw, circle, above=1cm of B] (C) {C};
  \node [draw, circle, below=1cm of B] (D) {D};
  
  \draw[-latex] (A) to [out=15, in=165] (B);
  \draw[-latex] (B) to [out=195, in=345] (A);
  
  \draw[-latex] (B) to [out=105, in=255] (C);
  \draw[-latex] (C) to [out=285, in=75] (B);
  
  \draw[-latex] (B) to [out=285, in=75] (D);
  \draw[-latex] (D) to [out=105, in=255] (B);
\end{tikzpicture}
\vspace{-0.5cm}
}

%% file: illustrations/dual_graph.tikz.tex
\usetikzlibrary{shapes,arrows,positioning,fit,backgrounds}
\scalebox{0.75}{
\begin{tikzpicture}[align=center]
  \node [draw, circle] (AB) {AB};
  \node [draw, circle, below=1cm of AB] (BA) {BA};
  
  \node [draw, circle, right=1cm of AB] (BC) {BC};
  \node [draw, circle, right=1cm of BC] (CB) {CB};

  \node [draw, circle, right=1cm of BA] (DB) {DB};
  \node [draw, circle, right=1cm of DB] (BD) {BD};
  
  \draw[-latex] (AB) to [out=285, in=75] (BA);
  \draw[-latex] (BA) to [out=105, in=255] (AB);
  \draw[-latex] (AB) -- (BD);
  \draw[-latex] (AB) -- (BC);

  \draw[-latex] (BC) to [out=15, in=165] (CB);
  \draw[-latex] (CB) to [out=195, in=345] (BC);

  \draw[-latex] (DB) to [out=15, in=165] (BD);
  \draw[-latex] (BD) to [out=195, in=345] (DB);

  \draw[-latex] (CB) -- (BA);
  \draw[-latex] (CB) -- (BD);

  \draw[-latex] (DB) -- (BA);
  \draw[-latex] (DB) -- (BC);

  
  
\end{tikzpicture}
\vspace{-0.5cm}
}

%% file: method.tex
\section{Proposed Method}\label{sec:method}
The \emph{\glsfirst{rfn}} aims to address the shortcomings of state-of-the-art \glspl{gcn} in the context of machine learning on road networks. The basic premise is to learn representations based on two distinct, but interdependent views: the node-\REVISED{relational}\REVSPACE{} and edge-relational views.

\subsection{Node-Relational and Edge-Relational Views}
In the node-relational view, we seek to learn representations of nodes, i.e., intersections, based on their node attributes and the relationships between nodes indicated by the edges $E$ in the primal graph representation of a road network $G^P = (V, E)$ and described by their edge attributes.
Similarly, we seek to learn representations of edges, i.e., road segments, in the edge-relational view, based on their edge attributes and the relationships between edges indicated by the between-edges $B$ in the dual graph representation of a road network $G^D = (E, B)$.
The relationship between two adjacent roads $(u, v)$ and $(v, w)$ is described by the attributes of the between-edge connecting them in the dual graph, including the angle between them, but also the attributes of the node $v$ that connects them.

The node-\REVISED{relational}\REVSPACE{} and edge-relational views are complementary.
The representation of a node in the node-relational view is dependent on the representation of the edges to its neighbors.
Similarly, the representation of an edge in the edge-relational view is dependent on the representation of the nodes that it shares with its neighboring edges.
Finally, the representation of an edge is also dependent on the representation of the between-edge connecting them in the dual graph.
\glspl{rfn} can exploit these two complementary views to leverage node, edge, and between-edge attributes simultaneously.


\subsection{Method Overview}
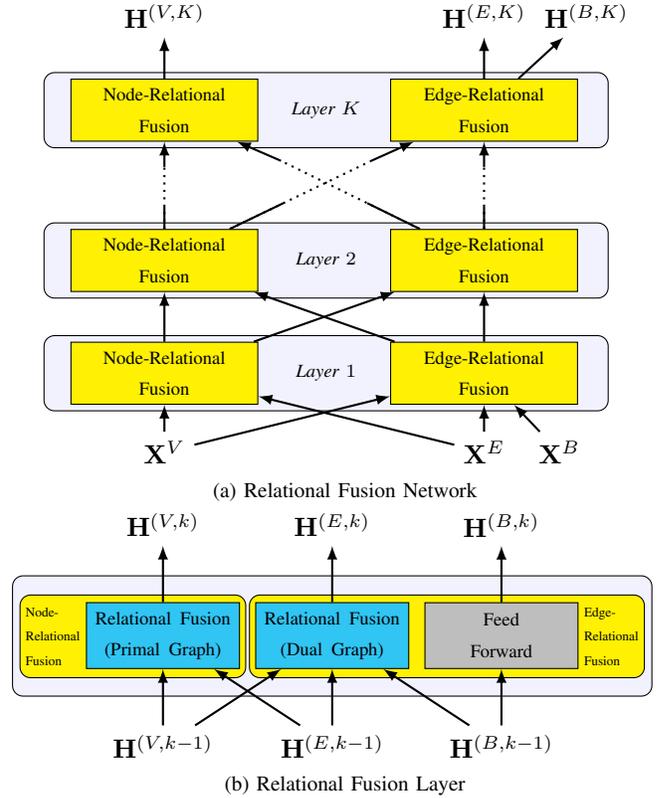
\begin{figure}[ht]
  \vspace*{-0.35cm}
  \centering
  \begin{subfigure}{\columnwidth}
    \centering
    \input{illustrations/overview.tikz.tex}
    \vspace*{0.4cm}
    \caption{Relational Fusion Network\label{fig:methodOverviewA}}
  \end{subfigure}
  \begin{subfigure}{\columnwidth}
    \centering
    \input{illustrations/layer.tikz.tex}
    \vspace*{-0.125cm}
    \caption{Relational Fusion Layer\label{fig:methodOverviewB}}
  \vspace*{-0.1cm}
  \end{subfigure}
  \caption{Overview of our method showing (a) a $K$-layered relational fusion network and (b) a relational fusion layer.\label{fig:methodOverview}}
  \vspace*{-0.5cm}
\end{figure}

\cref{fig:methodOverviewA} gives an overview of our method.
As shown, an \gls{rfn} consists of $K$ relational fusion layers, where $K \geq 1$.
It takes as input feature matrices $\mathbf{X}^V \in \mathbb{R}^{|V| \times d_V}$, $\mathbf{X}^E \in \mathbb{R}^{|E| \times d_E}$, and $\mathbf{X}^B \in \mathbb{R}^{|B| \times d_B}$ that numerically encode the node, edge, and between-edge attributes, respectively.
These inputs are propagated through each layer.
Each of these relational fusion layers, performs \emph{node-relational fusion} and \emph{edge-relational fusion} to learn representations from the node-\REVISED{relational}\REVSPACE{} and edge-relational views, respectively. 

Node-relational fusion is carried out by performing relational fusion on the primal graph.
Relational fusion is a novel graph convolutional operator that we describe in detail in \cref{sec:relationalFusion}.
In brief, relational fusion computes relational representations for, e.g., each relation $(v, n)$ of a node $v$ in the primal graph.
These relational representations are a fusion of the node representations of $v$ and $n$, but also the edge representation of $(v, n)$.
Finally, the relational representations are aggregated into a new representation of $v$.
Relational fusion differs from regular graph convolution by replacing the aggregate over neighbor representations with an aggregate over relational representations.

Edge-relational fusion is performed similarly to node-relational fusion but is applied on the dual graph representation.
Recall, that in the edge-relational view, the relation between two edges $(u, v)$ and $(v, w)$ is in part described by their between-edge attributes, but also by the node attributes of $v$ that describes the characteristics of the intersection between them.
Thus, as illustrated in \cref{fig:methodOverviewB}, relational fusion on the dual graph requires both node and between-edge information to compute relational aggregates.

The interdependence between the node-\REVISED{relational}\REVSPACE{} and edge-relational views is captured by using the node and edge representations from the previous layer $k-1$ as input to node-relational and edge-relational fusion in the next layer $k$ as illustrated by \cref{fig:methodOverview}.
Therefore each layer also applies regular feed-forward propagation on the between-edge representations to learn more abstract between-edge representations as input to the next layer.
Unlike node-relational and edge-relational fusion, this feed-forward propagation does not consider neighborhood information.

\cref{fig:methodOverviewB} gives a more detailed view of a relational fusion layer.
Each layer $k$ takes as input the learned node, edge, and between-edge representations from layer $k-1$, denoted by $\mathbf{H}^{(V, k-1)}$, $\mathbf{H}^{(E, k-1)}$, and $\mathbf{H}^{(B, k-1)}$, respectively.
Then node-relational and edge-relational fusion are performed to output new node, edge, and between-edge representations $\mathbf{H}^{(V, k)}$, $\mathbf{H}^{(E, k)}$, and $\mathbf{H}^{(B, k)}$.

\subsection{Relational Fusion}\label{sec:relationalFusion}
We present the pseudocode for the relational fusion operator at the $k$th layer in \cref{alg:relationalFusion}.
For clarity of notation, we write the notation from the perspective of a primal graph, but it can be applied to any graph, including a dual graph.
The operator takes as input a graph $G = (V, E)$, that is either the primal or dual graph representation of a road network, along with appropriate feature matrices $\mathbf{H}^{(V, k-1)}$ and $\mathbf{H}^{(E, k-1)}$ to describe nodes and edges in $G$.
Then, a new representation is computed for each node $v \in V$ by first computing relational representations at line~$4$. Given a node $v$, each relation $(v, n) \in N(v)$ that $v$ participates in, is converted to a \emph{relational representation}.

\begin{algorithm}
  \caption{The Node-Relational Fusion Operator\label{alg:relationalFusion}}
  \begin{algorithmic}[1]\raggedright
    \scriptsize
    \Function{RelationalFusion$^k$}{$G=(V, E)$, $\mathbf{H}^{(V, k-1)}$, $\mathbf{H}^{(E, k-1)}$}
      \State{\textbf{let} $\mathbf{H}^{(V, k)}$ be an arbitrary $|V| \times d_{(F, k)}$ real feature matrix.}
      \ForAll{$v \in V$} 
          \State{$F_{v} \gets \big\{$ \par
          \hspace{\algorithmicindent{}} \hspace{\algorithmicindent{}} \hspace{\algorithmicindent{}}
          $\textsc{Fuse}^k(\mathbf{H}^{(V,k-1)}_{v}, \mathbf{H}^{(E, k-1)}_{(v, n)}, \mathbf{H}^{(V, k-1)}_{n}) 
                                     \mid n \in N(v)\big\}$}
          \State{$\mathbf{H}^{(V, k)}_{v} \gets \textsc{Aggregate}^k(F_{v})$}
          \State{$\mathbf{H}^{(V, k)}_{v} \gets \textsc{Normalize}^k(\mathbf{H}^{(V, k)}_{v})$}
        \EndFor
        \State{\Return{$\mathbf{H}^{(V, k)}$}}
      \EndFunction
  \end{algorithmic}
\end{algorithm}

In \cref{alg:relationalFusion}, the relational representations at layer $k$ are computed by a fusion function $\textsc{Fuse}^k$ that outputs $d_{(F, k)}$-dimensional relational representations.
For each relation $(v, n)$, $\textsc{Fuse}^k$ takes as input representations of the source $v$ and target $n$ of the relation, $\mathbf{H}^{(V, k-1)}_{v}$ and $\mathbf{H}^{(V, k-1)}_{n}$, respectively, along with a representation $\mathbf{H}^{(E, k-1)}_{(v, n)}$ describing their relation, and fuses them into a relational representation of dimensionality $d_{(F, k)}$.
The relational representations are subsequently fed to a $\textsc{Aggregate}^k$ function, that aggregates them into a single $d_o$-dimensional representation of $v'$ of dimensionality.
Finally, the representation of $v$ may optionally be normalized by invoking the $\textsc{Normalize}^k$ function, e.g., using $L_2$ normalization~\citep{graphsage}.
The normalization step is particularly important if the relational aggregate has different scales across elements with different neighborhood sizes, e.g., if the aggregate is a sum.

Next, we discuss different choices for the fuse functions $\textsc{Fuse}^k$ and the relational aggregators $\textsc{Aggregate}^k$.

\subsection{Fusion Functions}
The fusion function is responsible for extracting useful information from each relation, thereby allowing an \gls{rfn} to create sharp boundaries at the edges of homophilic regions.
The fusion function thus plays an important role w.r.t.\ capturing volatile homophily.
We propose two fusion functions: $\textsc{AdditiveFuse}$ and $\textsc{InteractionalFuse}$.

\textsc{AdditiveFuse} takes as input source, target, and relation representations $\mathbf{H}^{(V, k-1)}_{v}$, $\mathbf{H}^{(V, k-1)}_{n}$, and $\mathbf{H}^{(E, k-1)}_{(v, n)}$, and transforms these sources:
\begin{multline}\label{eq:additiveFuse}
  \textsc{AdditiveFuse}^k(\mathbf{H}^{(V, k-1)}_{v}, \mathbf{H}^{(V, k-1)}_{n}, \mathbf{H}^{(E, k-1)}_{(v, n)}) = \\
  \sigma\big(\mathbf{H}^{(R, k-1)}_{(v, n)}\mathbf{W}^R + \mathbf{b}\big),
\end{multline}
where the relational feature matrix $\mathbf{H}^{(R, k-1)}_{(v, n)} \in \mathbb{R}^{d_R}$ is the vector concatenation of $\mathbf{H}^{(V, k-1)}_{v}$, $\mathbf{H}^{(V, k-1)}_{n}$, and $\mathbf{H}^{(E, k-1)}_{(v, n)}$.
Next, $\sigma$ is an activation function, $\mathbf{W}^R \in \mathbb{R}^{d_R \times d_o}$ is a weight matrix, $\mathbf{b} \in \mathbb{R}^{1 \times d_o}$ is a bias term, and $d_o$ is the output dimensionality of the fusion function (and its parent relational fusion operator).
\cref{eq:additiveFuse} is additive in the sense that it is equivalent to transforming the source, target, and relational representations with independent weight matrices and summing them before applying activation function $\sigma$.

$\textsc{AdditiveFuse}$ summarizes the relationship between $v$ and $n$, but does not explicitly model interactions between representations.
Hence, we propose an interactional fusion function:
\begin{multline}
  \textsc{InteractionalFuse}^k(\mathbf{H}^{(V, k-1)}_{v}, \mathbf{H}^{(V, k-1)}_{n}, \mathbf{H}^{(E, k-1)}_{(v, n)}) = \\
  \sigma\big((\mathbf{H}^{(R, k-1)}\mathbf{W}^I \odot \mathbf{H}^{(R, k-1)})\mathbf{W}^R\big) + \mathbf{b},
\end{multline}
where $\mathbf{W}^I \in \mathbb{R}^{d_R \times d_R}$ is a trainable interaction weight matrix, $\odot$ denotes element-wise multiplication, and $\mathbf{b} \in \mathbb{R}^{1 \times d_o}$ is a bias term. 
Notice that \textsc{InteractionalFuse} has a quadratic form. When computing the term $\mathbf{H}^{(I, k)} = (\mathbf{H}^{(R, k-1)}\mathbf{W}^I) \odot \mathbf{H}^{(R, k-1)}$, the $i$th ($1 \leq i \leq d_R$) value of vector $\mathbf{H}^{(I, k)}$ is
\REVISED{
\begin{equation}
  \mathbf{H}^{(I, k)}_i = \sum_{j=1}^{d_R} \mathbf{W}^I_{i, j}\mathbf{H}^{(R, k-1)}_i\mathbf{H}^{(R, k-1)}_j.
\end{equation}
}
In other words, $\mathbf{H}^{(I, k)}_i$ is the weighted sum of all interactions between the $i$th feature and all other features of the relational representation $\mathbf{H}^{(R, k-1)}$.
Modeling these interaction terms explicitly enables $\textsc{InteractionalFuse}$ to capture and weigh interactions at a much finer granularity than $\textsc{AdditiveFuse}$.

\textsc{InteractionalFuse} offers improved modeling capacity over \textsc{AdditiveFuse} to better address the challenge of volatile homophily, but at the cost of an increase in number of parameters that is quadratic in the number of input features.

\subsection{Aggregation Functions}
Different \textsc{Aggregate} functions have been proposed in the literature, and many are directly compatible with the relational fusion layer.
For instance, mean, and max/mean pooling aggregators~\citep{graphsage}.

Recently, aggregators based on attention mechanisms from the domain of natural language processing have appeared~\citep{gat}.
Such graph attention mechanisms allow a \gls{gcn} to filter out irrelevant or aberrant neighbors by weighing the contribution of each neighbor to the neighborhood aggregate.
This filtering property is highly desirable for road networks where even a single aberrant neighbor can contribute significant noise to an aggregate due to the low density of road networks.
In addition, it may help the network distinguish adjacent regions from each other and thereby improve predictive performance in the face of volatile homophily.

Current graph attention mechanisms rely on a common transformation of each neighbor thus rendering them incompatible with \glspl{rfn}, which rely on the context-dependent neighbor transformations performed by the \textsc{Fuse} function at each relational fusion layer.
We therefore propose an attentional aggregator that is compatible with our proposed \glspl{rfn}.


\paragraph{Attentional Aggregator}\label{sec:attentionalAggregator}
The attentional aggregator we propose computes a weighted mean over the relational representations.
Formally, the attentional aggregator computes the $\textsc{Aggregate}$ in \cref{alg:relationalFusion} as
\begin{equation}\label{eq:attentionalAggregator}
  \textsc{Aggregate}(F_v) = \sum_{\mathbf{f}_n \in F_v} A(v,n)\mathbf{f}_n,
\end{equation}
where $F_v = \{\mathbf{f}_n \mid n \in N(v)\}$ is the set of fused relational representations of node $v$'s relations to each of its neighbors $n \in N(v)$ computed at line $4$ in \cref{alg:relationalFusion}.
Furthermore, $A$ is an attention function, and $A(v, n)$ is the \emph{attention weight} that determines the contribution of each neighbor $n$ to the neighborhood aggregate of node $v$.

An attention weight $A(v, n)$ in \cref{eq:attentionalAggregator} depends on the relationship between a node $v$ and its neighbor $n$ in the input graph $G$ to \cref{alg:relationalFusion}.
This relationship is described by the relational feature matrix $\mathbf{H}^{(R, k-1)}_{(v, n)} \in \mathbb{R}^{d_R}$ and the attention weight is computed as
\begin{equation}\label{eq:attentionWeight}
  A(v, n) = \frac{
    \exp\big(C(\mathbf{H}^{(R, k-1)}_{(v, n)}) \big)}
  {\sum_{m \in N(v)} \exp\big( C(\mathbf{H}^{(R, k-1)}_{(v, m)}) \big)},
\end{equation}
where $C \colon \mathbb{R}^{d_R} \xrightarrow{} \mathbb{R}$ is an \emph{attention coefficient} function
\begin{equation}
  C(\mathbf{H}^{(R, k-1)}_{(v, n)}) =
      \sigma(\mathbf{H}^{(R, k-1)}_{(v, n)} \mathbf{W}^C),
\end{equation}
where $\mathbf{W^C} \in \mathbb{R}^{d_R}$ is a weight matrix and $\sigma$ is an activation function.
This corresponds to computing the softmax over the attention coefficients of all neighbors $n$ of $v$.
All attention weights sum to one, i.e., $\sum_{n \in N(v)} A(v, n) = 1$, thus making \cref{eq:attentionalAggregator} a weighted mean.
In other words, each neighbor's contribution to the mean is regulated by an attention weight, thus allowing an RFN to reduce the influence of (or completely ignore) aberrant neighbors that would otherwise contribute significant noise to the neighborhood aggregate.

\subsection{Forward Propagation}
\begin{algorithm}
  \caption{Forward Propagation Algorithm\label{alg:forward-propagation}}
  \scriptsize
  \begin{algorithmic}[1]\raggedright
    \Function{ForwardPropagation}{$\mathbf{X}^V$, $\mathbf{X}^E$, $\mathbf{X}^B$}
      \State{\textbf{let} $\mathbf{H}^{(V, 0)} = \mathbf{X}^V$, $\mathbf{H}^{(E, 0)} = \mathbf{X}^{E}$, and $\mathbf{H}^{(B, 0)} = \mathbf{X}^{B}$}
      \For{$k=1$ to $K$}
        \State{$\mathbf{H}^{(V, k)} \gets \textsc{RelationalFusion}^k(
          G^P, \mathbf{H}^{(V, k-1)}, \mathbf{H}^{(E, k-1)})$}
        \State{$\mathbf{H}^{(B', k-1)} \gets \textsc{Join}(\mathbf{H}^{(V, k-1)}, \mathbf{H}^{(B, k-1)})$}
        \State{$\mathbf{H}^{(E, k)} \gets \textsc{RelationalFusion}^k(
          G^D, \mathbf{H}^{(E, k-1)}, \mathbf{H}^{(B', k-1)})$}
        \State{$\mathbf{H}^{(B, k)} \gets \textsc{FeedForward}^k(\mathbf{H}^{(B, k-1)})$}
        \EndFor
        \State{\Return{$\mathbf{H}^{(V, K)}$, $\mathbf{H}^{(E, K)}$, $\mathbf{H}^{(B, K)}$}}
      \EndFunction
    \Function{Join}{$\mathbf{H}^{V}$, $\mathbf{H}^{E}$}
    \ForAll{$b \in B$}
      \State{\textbf{let} $b = \big( (u, v), (v, w) \big)$}
      \State{$\mathbf{H}^{B'}_{b} \gets \mathbf{H}^B_{b} \oplus \mathbf{H}^V_v$}
    \EndFor
    \State{\Return{$\mathbf{H}^{B'}$}}
    \EndFunction
  \end{algorithmic}
\end{algorithm}

With the relational fusion operator and its components defined in Sections 4.3 to 4.5, we proceed to explain how forward propagation is performed through the relational fusion layers (introduced in Section 4.2) of an \gls{rfn}. The forward propagation algorithm is shown in Algorithm 2.
Starting from the input encoding of node, edge, and between-edge attributes, each layer $k$ transforms the node, edge, and between-edge representations emitted from the previous layer.
The node representations are transformed using node-relational fusion.
This is done by invoking the $\textsc{RelationalFusion}$ function at line~$4$ with the primal graph, and the node and edge representations from the previous layer as input.

Edge-relational fusion is performed at lines~$5\text{-}6$ to transform the edge representations from the previous layer.
In line~$5$, the node and between-edge representations from the previous layer are joined using the \textsc{Join} function (defined at lines~$11\text{-}16$) to capture the information from both sources that describe the relationships between two edges.
On line~$6$, $\textsc{RelationalFusion}$ is invoked again, now with the dual graph (indicating relationships between edges), and the edge representations from the previous layer.
The last input is the joined node and between-edge representations also from the previous layer.
The between-edge representations are transformed using a single feed-forward operator at line~$7$.

The number of layers in an \gls{rfn} determines which nodes, edges, and between-edges influence the output representations.
Each layer aggregates information from the relations to first-order node and edge neighbors during edge- and node-relational fusion, respectively.
Thus, a $K$-layer \gls{rfn} aggregates information up to a distance $K$ from nodes in the primal graph and edges in the dual graph, respectively.

Once the input representations have been propagated through all the layers, the final node, edge, and between-edge representations are output on line~$9$.
These three outputs allows the relational fusion network to be jointly optimized for node, edge, and between-edge predictions, e.g., when the network is operating in a multi-task learning setting.
However, if only one or two outputs are desired, the superfluous operations in the last layer can be skipped to save computational resources.
For instance, propagation of node and between-edges can be skipped by replacing line~$4$ and line~$7$ with $\mathbf{H}^{(V, k)} = \mathbf{H}^{(V, k-1)}$ and $\mathbf{H}^{(B, k)} = \mathbf{H}^{(B, k-1)}$, respectively, when $k = K$.

%% file: illustrations/overview.tikz.tex
\usetikzlibrary{shapes,arrows,positioning,fit,backgrounds,calc}
\begin{tikzpicture}[
    align=center,
    layer/.style={shape=rectangle, text width=2.25cm, align=center, draw=black},
    arrow/.style={-latex, thick}
  ]
   \def\layerdistance{0.55cm}
   \def\layeroffset{1.75}
   \def\inoutoffset{0.25cm}

    \node[] (Xn) at (0.75,-2*\inoutoffset) {$\mathbf{X}^V$};
    \node[] (Xe) at (5,-2*\inoutoffset) {$\mathbf{X}^E$};
    \node[] (Xbe) at (6,-2*\inoutoffset) {$\mathbf{X}^{B}$};
   
    \node [draw, fill=blue!50!white!10, rounded corners, minimum width=7.5cm, minimum height=1cm, outer sep=0, inner sep=0] at (2.9,0.55) {\scriptsize{\textit{Layer $1$}}};
  
    \node[layer, fill=yellow] (GCN1n) at (0.75,1*\layerdistance) {\scriptsize{Node-Relational Fusion}};
    \draw[arrow] (Xn) -- (GCN1n);
    \draw[arrow] (Xe) -- (GCN1n);
    
    \node[layer, fill=yellow] (GCN1e) at (5,1*\layerdistance) {\scriptsize{Edge-Relational Fusion}};
    \draw[arrow] (Xn) -- (GCN1e);
    \draw[arrow] (Xe) -- (GCN1e);
    \draw[arrow] (Xbe) -- (GCN1e);

    \node [draw, fill=blue!50!white!10, rounded corners, minimum width=7.5cm, minimum height=1cm, outer sep=0, inner sep=0] at (2.9,2.05) {\scriptsize{\textit{Layer $2$}}};

    \node[layer, fill=yellow] (GCN2n) at (0.75, 1.5cm + 1*\layerdistance) {\scriptsize{Node-Relational Fusion}};
    \draw[arrow] (GCN1n) -- (GCN2n);
    \draw[arrow] (GCN1e) -- (GCN2n);
    
    \node[layer, fill=yellow] (GCN2e) at (5, 1.5cm + 1*\layerdistance) {\scriptsize{Edge-Relational Fusion}};
    \draw[arrow] (GCN1e) -- (GCN2e);
    \draw[arrow] (GCN1n) -- (GCN2e);
   
    \node [draw, fill=blue!50!white!10, rounded corners, minimum width=7.5cm, minimum height=1cm, outer sep=0, inner sep=0] (LK) at (2.9,4.05) {\scriptsize{\textit{Layer $K$}}};

    \node[layer, fill=yellow] (GCNkn) at (0.75,3.5cm + 1*\layerdistance) {\scriptsize{Node-Relational Fusion}};
    \node[draw=none] (A3) at (2.3, 3.5) {};
    \draw[arrow] (A3.south) -- (GCNkn);
    \draw[thick] (GCN2e) edge[] ($(GCN2e)!0.575!(A3.south)$) edge [dotted] ($(GCN2e)!1.0!(A3.south)$);
    
    \node[draw=none, above=0.825cm of GCN2n] (A1) {};
    \draw[arrow] (A1.south) -- (GCNkn);
    \draw[thick] (GCN2n) edge[] ($(GCN2n)!0.575!(A1.south)$) edge [dotted] ($(GCN2n)!1.0!(A1.south)$);
    
    \node[layer, fill=yellow] (GCNke) at (5, 3.5cm + 1*\layerdistance) {\scriptsize{Edge-Relational Fusion}};
    \node[draw=none, above=0.825cm of GCN2e] (A2) {};
    \draw[arrow] (A2.south) -- (GCNke);
    \draw[thick] (GCN2e) edge[] ($(GCN2e)!0.575!(A2.south)$) edge [dotted] ($(GCN2e)!1.0!(A2.south)$);

    \node[draw=none] (A4) at (3.5, 3.5) {};
    \draw[arrow] (A4.south) -- (GCNke);
    \draw[thick] (GCN2n) edge[] ($(GCN2n)!0.575!(A4.south)$) edge [dotted] ($(GCN2n)!1.0!(A4.south)$);
    \node[above=\layerdistance of GCNkn] (Zn) {$\mathbf{H}^{(V, K)}$};
    \draw[arrow] (GCNkn) -- (Zn);
    
    \node[above=\layerdistance of GCNke] (Ze) {$\mathbf{H}^{(E, K)}$};
    \draw[arrow] (GCNke) -- (Ze);
    
    \node[right=0cm of Ze] (Zb) {$\mathbf{H}^{(B, K)}$};
    \draw[arrow] (GCNke) -- (Zb);
    
\end{tikzpicture}
\vspace{-0.5cm}

%% file: illustrations/layer.tikz.tex
\usetikzlibrary{shapes,arrows,positioning,fit,backgrounds}
\begin{tikzpicture}[
    align=center,
    layer/.style={shape=rectangle, text width=1.8cm, align=center, draw=black, minimum height=0.875cm},
    arrow/.style={-latex, thick}
  ]
    \node [draw, fill=blue!50!white!10, rounded corners, minimum width=8.5cm, minimum height=1.6cm, outer sep=0, inner sep=0] at (2.25,1.22) {};
    
    \node [draw, fill=yellow, rounded corners, minimum width=3cm, minimum height=1.1cm, outer sep=0, inner sep=0] at (-0.4,1.215) {};
    \node [opacity=0, text opacity=1, outer sep=0, inner sep=0, text width=2.5cm, align=left] at (-0.575,1.215) {\tiny{Node-\vspace{-0.1cm}\break Relational\vspace{-0.1cm}\break Fusion}};
  
    \node [draw, fill=yellow, rounded corners, minimum width=5.25cm, minimum height=1.1cm, outer sep=0, inner sep=0] at (3.775,1.215) {};
    \node [opacity=0, text opacity=1, outer sep=0, inner sep=0, text width=1cm, align=left] at (6.1,1.215) {\tiny{Edge-\vspace{-0.1cm}\break Relational\vspace{-0.1cm}\break Fusion}};

   \def\layerdistance{0.5cm}
   \def\inoutoffset{0.25cm}
    \node[] (Xn) at (0,-\inoutoffset) {$\mathbf{H}^{(V, k-1)}$};
    \node[] (Xe) at (2.25,-\inoutoffset) {$\mathbf{H}^{(E, k-1)}$};
    \node[] (Xbe) at (4.5,-\inoutoffset) {$\mathbf{H}^{(B, k-1)}$};

    \node[layer, above=\layerdistance + \inoutoffset of Xn, fill=cyan!255!white!25] (GCN1n) {\scriptsize{Relational Fusion\\(Primal Graph)}};
    \draw[arrow] (Xn) -- (GCN1n);
    \draw[arrow] (Xe) -- (GCN1n);
    
    \node[layer, above=\layerdistance + \inoutoffset of Xe, fill=cyan!255!white!25] (GCN1e) {\scriptsize{Relational Fusion\\(Dual Graph)}};
    \draw[arrow] (Xn) -- (GCN1e);
    \draw[arrow] (Xe) -- (GCN1e);
    \draw[arrow] (Xbe) -- (GCN1e);

    \node[layer, above=\layerdistance + \inoutoffset of Xbe, fill=gray!255!white!20] (FF1be) {\scriptsize{Feed\\Forward}};
    \draw[arrow] (Xbe) -- (FF1be);
    
    \node[above=\layerdistance + \inoutoffset of GCN1n] (Zn) {$\mathbf{H}^{(V, k)}$};
    \draw[arrow] (GCN1n) -- (Zn);
    
    \node[above=\layerdistance + \inoutoffset of GCN1e] (Ze) {$\mathbf{H}^{(E, k)}$};
    \draw[arrow] (GCN1e) -- (Ze);
    
    \node[above=\layerdistance + \inoutoffset of FF1be] (Zbe) {$\mathbf{H}^{(B, k)}$};
    \draw[arrow] (FF1be) -- (Zbe);
\end{tikzpicture}

%% file: experiments_setup.tex
\section{Experimental Evaluation}\label{sec:experiments}
We evaluate our method on two machine learning tasks: driving speed estimation and speed limit classification.
These tasks represent a regression task and a classification task, respectively. We evaluate our method on both within-network prediction and cross-network prediction.

In the within-network setting, models are trained and evaluated on the road network of the same municipality.
In the cross-network setting, models are trained on the road network of one municipality, but tested on another, unseen road network.
For all tasks and settings, we compare the results of our method against a number \REVISED{of}\REVSPACE{} baselines, including state-of-the-art \glspl{gcn}.
Finally, we conduct a case study to investigate the behavior of \glspl{rfn} under conditions of volatile homophily.

\subsection{Data Set}\label{sec:data-set}
\begin{table}
  \centering
  \caption{Road network sizes.\label{tab:network-sizes}}
  \begin{tabular}{llll}
    \toprule
    & \emph{AAL} & \emph{BRS} & \emph{CPH} \\
    \midrule
    No.\ of Nodes ($|V|$) & $16\,294$ & $5\,889$ & $10\,738$ \\
    No.\ of Edges ($|E|$) & $35\,947$ & $13\,073$ & $26\,117$ \\
    No.\ of Between-Edges ($|B|$) & $94\,718$ & $34\,428$ & $72\,147$ \\
    \bottomrule
  \end{tabular}
   \vspace*{-0.15cm}
\end{table}
We extract the spatial representation of the Danish municipalities of Aalborg (AAL), Brønderslev (BRS), and Copenhagen (CPH) from \gls{osm}~\citep{osm} and convert them to their primal and dual graph representations as described in \cref{sec:preliminaries}.
The sizes of these networks can be seen in \cref{tab:network-sizes}. 
We combine the \gls{osm} data with a zone map from the Danish Business Authority\footnote{\url{https://danishbusinessauthority.dk/plansystemdk}}.
From these two data sources, we derive $3$ node features, $16$ edge features, and $5$ between-edge features. See Appendix~\ref{app:feature-derivation} for further details on the feature derivation process.

For the driving speed estimation task, we use a dataset of observed driving speeds, each matched to a road segment, derived from a set of vehicle trajectories \REVISED{collected between January 1 2012 and December 31 2014}~\citep{driving_speeds}.
Each such observed driving speed corresponds to a traversal in this set of trajectories~\citep{driving_speeds}.
\REVISED{Further details on the vehicle trajectory data can be found elsewhere~\citep{driving_speeds}.}

For speed limit classification, we use $19\,510$ speed limits collected from the \gls{osm} data and additional speed limits collected from the municipality of Aalborg. This dataset is highly imbalanced, e.g., some speed limits are several thousands of times more frequent than others.

We split speed limits and driving speeds into training, validation, and test sets. For the driving speeds, the split is temporal s.t.\ the oldest observations are in the training set, the newest are in the test set, and the driving speeds in the validation set cover a period in-between.
\cref{tab:data-set-sizes} shows the total number of driving speeds and speed limits with the proportion between training, validation, and testing sets.
 
 \begin{table}
   \caption{Dataset sizes and splits.\label{tab:data-set-sizes}}
   \centering
  \begin{tabular}{lll}
    \toprule
    & \emph{Size} & \emph{Train / Val / Test} \\
    \midrule 
    Driving Speeds (AAL) & $8\,675\,599$ & $0.46$ / $0.21$ / $0.32$\\
    Driving Speeds (BRS) & $847\,963$ & $0.00$ / $0.00$ / $1.00$\\
    \midrule
    Speed Limits (AAL)  & $19\,510$ & $0.50$ / $0.23$ / $0.27$\\
    Speed Limits (CPH) & $17\,824$ & $0.00$ / $0.00$ / $1.00$\\
    \bottomrule
  \end{tabular}
   \vspace*{-0.15cm}
 \end{table}

\subsection{Algorithms}
We use the following \gls{gcn} baselines for comparison in the experiments: (1) the Max-Pooling variant of GraphSAGE~\citep{graphsage} and the (2) Graph Attention Network~\citep{gat}. In addition, we include two non-\gls{gcn} baselines: (3) a regular Multi-Layer Perceptron (MLP), and (4) a Grouping Estimator.
The Grouping Estimator is a simple baseline that groups all road segments depending on their road category and on whether they are within a city zone or not. At prediction time, the algorithm outputs the mean (for regression) or mode (for classification) label of the group a particular road segment belongs to.

In our experiments, we include four variants of the \gls{rfn} that combines different relational aggregators and fusion functions.
We examine attentional (A) aggregation and non-attentional (N) aggregation together with additive (A) and interactional fusion (I).
The non-attentional computes an unweighted mean over the relational representations, i.e., all neighbors have the same attention weight.
Each variant is denoted by combination their aggregator and fusion function acronyms, e.g., RFN-A+I.

\subsection{Experimental Setup}\label{sec:experimental-setup}
The GraphSAGE and GAT models are run on the dual graph representations of the road network s.t.\ they learn edge representations directly.
All models are two-layer models and use the ELU~\citep{elu} activation function on the first layer.
ReLU~\citep{relu} and softmax are used on the last layer for driving speed estimation and speed limit classification, respectively.
The ReLU~\citep{relu} activation function is used in the GraphSAGE pooling operation for both tasks.
We select layer sizes, learning rates, and the number of GAT attention heads by evaluating different hyperparameter configurations on the validation sets in a grid search and selecting the best-performing configuration.
Each configuration is repeated ten times, and the configuration with the highest mean performance is selected for final evaluation on the test set.
Appendix~\ref{app:hyperparam-selection} provide additional details on the hyperparameter selection process and documents the hyperparameters are used for each model in our experiments.

All neural network algorithms are implemented using the MXNet\footnote{\url{https://mxnet.incubator.apache.org/}} deep learning library.
We make our implementation of the relational fusion networks publicly available\footnote{https://github.com/TobiasSkovgaardJepsen/relational-fusion-networks}.

\paragraph{Model Training and Evaluation}
We initialize the weights of all neural network models using Xavier initialization~\citep{xavier} and train the models using the ADAM optimizer~\citep{adam} in (mini-)batches of $256$ segments. For training efficiency, the batches are pre-computed in a stratified manner s.t.\ the distributions of road categories and speed limits---for speed limit classification and driving speed estimation, respectively---approximate the distributions of the full training set.
The mini-batches are shuffled after each epoch.

We train all models for $20$ and $30$ epochs for driving speed estimation and speed limit classification, respectively, on Aalborg.
For the cross-network setting, we use Brønderslev and Copenhagen for driving speed estimation and speed limit classification, respectively.
For speed limit classification, the training set is randomly oversampled to address the class imbalance and regularize the models with early stopping.

For speed limit classification, we train the models using the binary cross entropy loss and evaluate the models using the $F_1$ macro score.
For driving speed estimation, we train the models using a per-segment mean squared loss $\sum_{y \in Y} \frac{(\hat{y} - y)^2}{|Y|}$, where $Y$ is the set of observed driving speeds associated with the segment, and $\hat{y}$ is the model's estimate of the driving speed of segment.
The per-batch loss is the mean per-segment loss.

Driving speed estimation models are evaluated using a per-segment \gls{mae}: $|\hat{y} - \bar{Y}|$, where $\bar{Y}=\sum_{y \in Y} \frac{y}{|Y|}$ is the mean recorded speed for the segment.
The final error of a model is the mean over all the per-segment errors.
However, the value of $\bar{Y}$ is highly sensitive to outliers.
We therefore exclude segments with fewer than ten observations when calculating the final error.

Using the per-segment losses and errors to train and evaluate models, respectively, on the driving speed estimation task avoids a bias towards models that perform well on popular road segments in the data set.

%% file: experiments_results.tex
\subsection{Results}
We train ten models for each algorithm and task on Aalborg Municipality experiments for each model.
We use these models for both within-\REVISED{network}\REVSPACE{} and cross-network inference and report the mean performance with standard deviations in \cref{tab:results}.
When reading \cref{tab:results}, note that low values and high values are desirable for driving speed estimation and speed limit classification, respectively.

The GAT algorithm can become unstable during training~\citep{gat} and we observed this phenomenon on one out of the ten runs on the driving speed estimation task.
The algorithm did not converge on this run and is therefore excluded from the results shown in \cref{tab:results}.
In addition, none of the models give reasonable results on cross-network speed limit classification where top-performance across all models is reduced to almost a third.
Hence, we also exclude these results from \cref{tab:results}.

\begin{table}[]
  \centering
  \caption{Algorithm performance on Driving Speed Estimation (DSE) and Speed Limit Classification (SLC) on AAL and BRS.\label{tab:results}}
  \begin{tabular}{lllll}
    \toprule
                                   & \multicolumn{2}{c}{\emph{DSE}} & \multicolumn{1}{c}{SLC} \\
    \cmidrule(lr){2-3} \cmidrule(lr){4-4}
    \textit{Algorithm}             & \multicolumn{1}{c}{\emph{AAL}}  & \multicolumn{1}{c}{\emph{BRS}} & \multicolumn{1}{c}{\emph{AAL}} \\
    \midrule
    GP                             & $11.026$           & $12.715$           & $0.356$           \\
    MLP                            & $10.160 \pm 0.119$ & $12.659 \pm 0.265$ & $0.443 \pm 0.027$ \\
    GAT                            & $9.548 \pm 0.151$  & $12.119 \pm 0.342$ & $0.442 \pm 0.018$ \\
    GraphSAGE                      & $8.960 \pm 0.115$  & $11.292 \pm 0.293$ & $0.432 \pm 0.014$ \\
    \midrule
    RFN-N+A                        & $7.685 \pm 0.189$          & $9.382 \pm 0.447$          & $0.500 \pm 0.011$          \\
    RFN-A+A                        & $7.440 \pm 0.133$          & $9.078 \pm 0.080$          & $0.518 \pm 0.022$          \\
    RFN-N+I                        & $6.911 \pm 0.080$          & $8.823 \pm 0.196$          & $0.507 \pm 0.012$          \\
    RFN-A+I                        & $\mathbf{6.797 \pm 0.124}$ & $\mathbf{8.587 \pm 0.238}$ & $\mathbf{0.535 \pm 0.014}$ \\
    \bottomrule
  \end{tabular}
\end{table}

\subsubsection{Within-Network Inference}\label{sec:baseline-comparison}
As shown in \cref{tab:results}, the \gls{rfn} variants outperform all baselines on both within-network driving speed estimation and within-network speed limit classification.
The best \gls{rfn} variant outperforms the state-of-the-art graph convolutional approaches, i.e., GraphSAGE and GAT, by $32\%$ and $40\%$, respectively, on the driving speed estimation task.
On the speed limit classification task, the best \gls{rfn} variant outperforms GraphSAGE and GAT by $24\%$ and $21\%$, respectively.

\cref{tab:results} shows that the \gls{rfn} variants achieve similar performance, but the attentional variants are superior to their non-attentional counterparts that use the same fusion function. In addition, the interactional fusion function appears to be strictly better than the additive fusion function.
Interestingly, GraphSAGE and GAT fail to outperform MLP on the speed limit classification task.
The MLP classifies road segments independent of any adjacent road segments.
Unlike the \gls{rfn} variants, it appears that GraphSAGE and GAT are unable to effectively leverage the information from adjacent road segments.
This supports our discussion in \cref{sec:introduction} of the problems with direct inheritance during neighborhood aggregation in the context of road networks.

\begin{figure*}[ht]
  \centering
  \scriptsize
  \setcounter{subfigure}{0}

  \definecolor{transportation}{RGB}{255, 127, 14}
  \definecolor{transition}{RGB}{44, 160, 44}
  \definecolor{residential}{RGB}{23, 190, 207}
  
  \begin{tabular}{p{0.3865\textwidth}ccccc}
    & \parbox{0.15\textwidth}{\centering\cellcolor{transportation} \textbf{Transportation}} & &
    \parbox{0.15\textwidth}{\centering\cellcolor{transition} \textbf{Transition}} & &
    \parbox{0.15\textwidth}{\centering\cellcolor{residential} \textbf{Residential}} \\
  \end{tabular}
  \begin{tabular}{ccc}
    \multirow[c]{1}{*}{\vspace*{0.35cm}\includegraphics[height=6.12cm, width=0.315\textwidth, trim={23cm, 0cm, 25cm, 0cm}, clip]{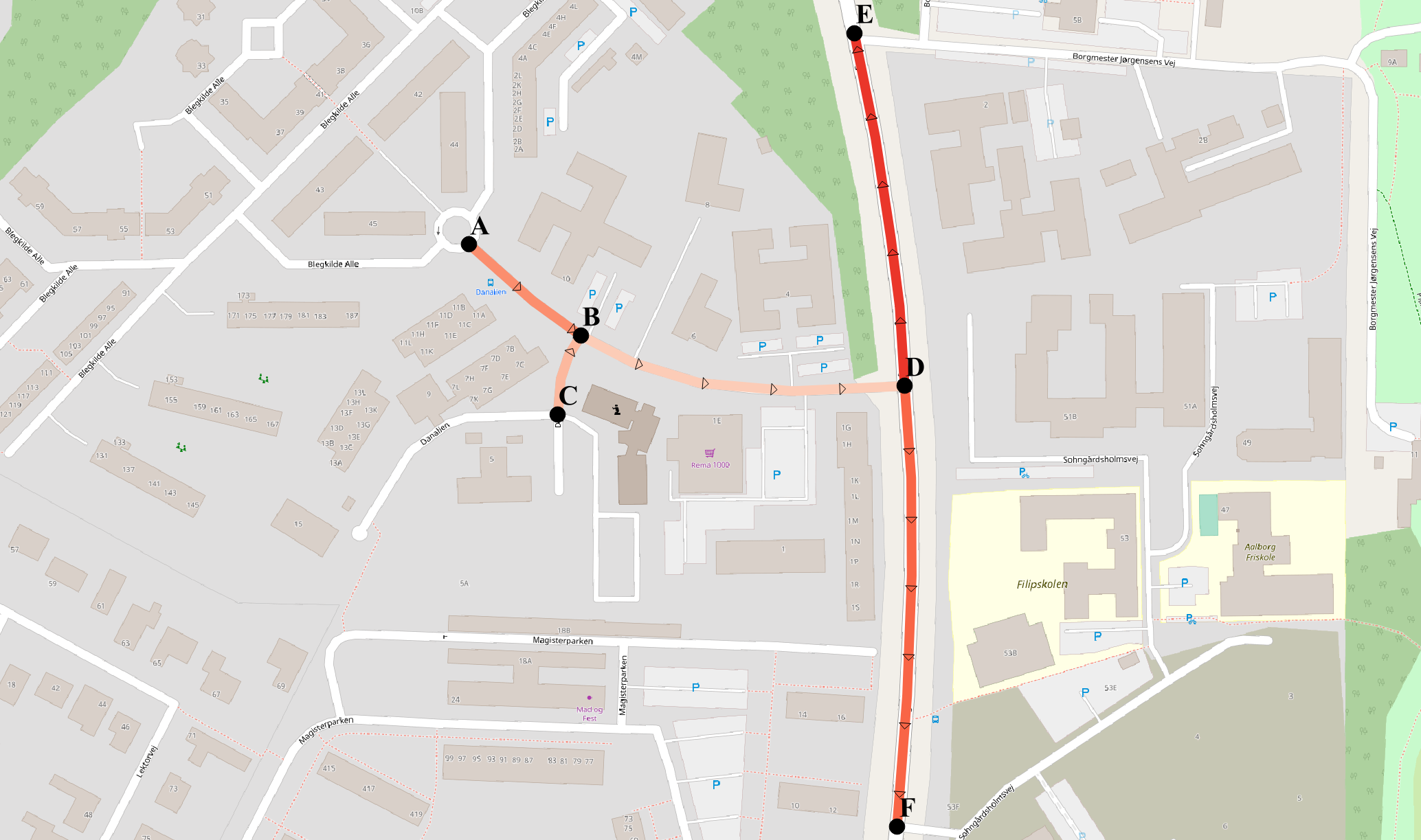}} \\
      & \includegraphics[width=0.295\textwidth, trim={1.1cm, 0.75cm, 1.475cm, 1.225cm}, clip]{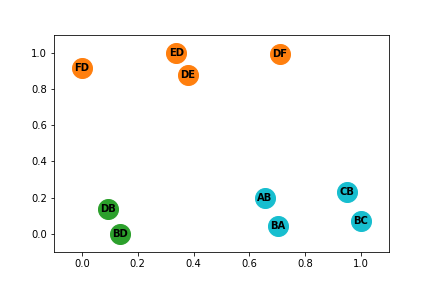} &
    \includegraphics[width=0.295\textwidth, trim={1.1cm, 0.75cm, 1.475cm, 1.225cm}, clip]{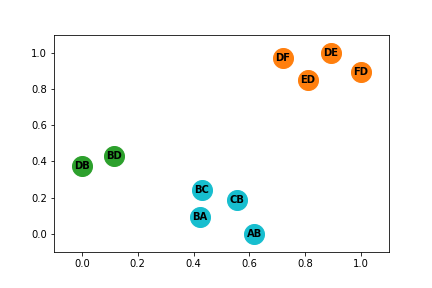} \\
    \stepcounter{figure}
    \stepcounter{figure}
     & \parbox{0.33cm + \widthof{\scriptsize{Input Representations}}}{\captionof{subfigure}{\scriptsize{Input Representations}}\label{fig:aggregations-input}} &
       \parbox{0.33cm + 0.005cm + \widthof{\scriptsize{RFN-A+I}}}{\captionof{subfigure}{\scriptsize{{RFN-A+I}}}\label{fig:aggregations-rfnai}} \\[0.2cm]
     & \includegraphics[width=0.295\textwidth, trim={1.1cm, 0.75cm, 1.475cm, 1.225cm}, clip]{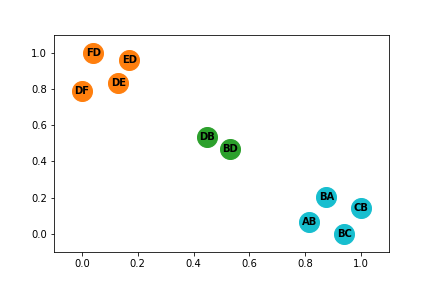} &
    \includegraphics[width=0.295\textwidth, trim={1.1cm, 0.75cm, 1.475cm, 1.225cm}, clip]{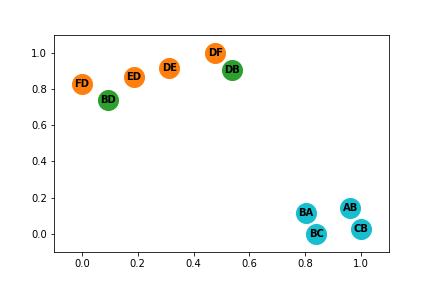} \\
     & \parbox{0.33cm + \widthof{\scriptsize{GraphSAGE}}}{\captionof{subfigure}{\scriptsize{GraphSAGE}}\label{fig:aggregations-graphsage}} & \parbox{0.33cm + 0.005cm + \widthof{\scriptsize{GAT}}}{\captionof{subfigure}{\scriptsize{GAT}\label{fig:aggregations-gat}}} \\[0.1cm]
    \setcounter{figure}{\value{figure}-2}
    \parbox[t]{0.35\textwidth}{\captionof{figure}{Ground truth driving speeds of the Danalien region in Aalborg, Denmark. Triangles indicate direction, black dots mark nodes, and color indicates speed: the darker, the faster.\label{fig:volatile-homophily-a}}} &
    \multicolumn{2}{c}{\parbox[t]{0.59\textwidth}{\captionof{figure}{\glsname{tsne} visualizations of the edge representations in the Danalien case (a) before and (b, c, d) after the first graph convolution in different models.\label{fig:aggregations}}}}\\
  \end{tabular}
  \setcounter{subfigure}{0}
  \vspace*{-0.35cm}
 \end{figure*}

\subsubsection{Cross-Network Inference}
\cref{tab:results} shows that the ranking of the models remain the same on driving speed estimation in the cross-network setting: all \gls{rfn} variants outperform the baselines and the RFN-A+I remains the best-performing variant.
All models suffer a performance loss, but the performance loss of the \gls{rfn} variants is smaller than that of the neural network baselines.
The best-performing baseline, GraphSAGE, has an error increase of $2.332$ km/h.
In comparison, the best-performing \gls{rfn} variant, RFN-A+I, has a $23\%$ smaller performance loss with an error increase of $1.790$ km/h.

The small performance loss of the \gls{rfn} variants compared to the neural network baselines on cross-network driving speed estimation suggests that (1) \glspl{rfn} may learn transferable knowledge that generalizes well to other road networks and (2) that they are robust to substantial changes to the road network structure.
However, we did not observe these tendencies on cross-network speed limit classification where all models achieved poor performance.
This suggests that knowledge transferability is highly task-dependent and that some tasks may require sophisticated transfer learning methods.

\subsection{Case Study: Danalien}\label{sec:danalien-case}
\REVISED{
\cref{tab:results} shows that \glspl{rfn} generally perform better than the GraphSAGE and GAT baselines. However, it is unclear whether \glspl{rfn} are capable of separating areas that are dissimilar, but internally homophilic, as intended. We have therefore examined the \gls{rfn}'s separation capabilities in areas exhibiting volatile homophily. For comparison, we have also examined the separation capabilities of the GraphSAGE and GAT algorithms.
For the sake of brevity, we present only our case study on the Danalien intersection that was introduced in \cref{fig:compelling-example}. In this study, we consider a larger part of the residential area, shown in \cref{fig:volatile-homophily-a}.
The Danalien case represents a quite typical scenario: a residential area that is connected to the rest of the city through a larger road. We have also examined a more extreme, but less typical case---cf.\ Appendix~\ref{app:dall-case}.
}

\COMMENT{
We examine the behavior of graph convolutions in \glspl{rfn} and state-of-the-art \glspl{gcn} under conditions of volatile homophily to gain insight into their differences.
To this end, we perform a case study on the Danalien intersection first introduced in \cref{fig:compelling-example}.
}

We select the \gls{rfn}, GraphSAGE, and GAT models with the best validation performance.
For these models, we compare the representations of the road segments in the Danalien case before and after applying the first graph convolutional layer by projecting these representations into two dimensions using \gls{tsne}~\citep{tsne} and normalizing the dimensions to be between $0$ and $1$. These representations are visualized in \cref{fig:aggregations}.

The segments AB, BA, BC, and CB in \cref{fig:volatile-homophily-a} form a residential area, and segments DE, ED, DF, and FD form a transportation area.
In addition, segments BD and DB are transition segments that must be traversed to transition between the two areas.
The residential, transportation, and transition segments are colored differently in the visualizations shown in \cref{fig:aggregations}.

Interestingly, the graph convolution in the different models have different effects on the relative position of the segments in the representation space.
In the input representation, illustrated in \cref{fig:aggregations-input}, the road segments are grouped with road segments of the same category.
As shown in \cref{fig:aggregations-rfnai}, the first graph convolution of the RFN-A+I model, makes these groups tighter and separates the residential and transportation segments further.
Although not included in \cref{fig:aggregations}, the other \gls{rfn} variants have a similar effect on the position of the road segments in the representation space.

As shown in \cref{fig:aggregations-graphsage}, the GraphSAGE model tightens the groups from \cref{fig:aggregations-input} like the RFN-A+I model, but places the transition segments roughly equi-distant from the residential and transportation clusters.
The transition segments are more similar in terms of driving speed to the residential segments, than the transportation segments
Therefore, this behavior of the GraphSAGE algorithm is expected since the GraphSAGE model inherits neighbor representations directly and indiscriminately (i.e., all neighbors contribute equally to the aggregate without an attention mechanism).
Since the transition segments have the same number of transportation and residential segments as neighbor segments, the representations of the transition segments is between the two other categories in the representation space.

The GAT algorithm functions in a very similar manner to the GraphSAGE algorithm, except that it can select which neighbor segments to inherit neighbor representations from using its attention mechanism.
The logic of this attention mechanism is shared across neighbors that has similar features (or belong to the same category in the Danalien example).
When computing a new representation for the transition segments, the GAT model will therefore give preference to the transportation segments, give preference to the residential segments, or give no preference at all.
As illustrated in \cref{fig:aggregations-gat}, the GAT model gives preference to the transportation segments both when computing representations for the transition segments and when computing representations for the transportation segments.
Thus, the GAT model groups the transition segments with the transportation segments, despite the transition segments being more similar to the residential segments in terms of driving speed.

As indicated by the tightness of the transition and transportation cluster in \cref{fig:aggregations-gat}, the preference for the transportation segments is substantial when computing transition segment representations.
We suspect that this behavior is attributed to two causes.
First, the relative rarity of transition segments causes poor predictive performance on such segments to have little impact on the total loss.
Second, road segments typically have at least one neighbor segment with a similar feature representation (e.g., belongs to the same category) and therefore such segments generally contribute little new information to distinguish a road segment of, e.g., the transition category from another road segment of the transition category.

In conclusion, we observe that the \gls{rfn} model is able to create separate the different segment categories in the Danalien case better than the GAT and GraphSAGE models are, which suggests that the \gls{rfn} architecture is more robust to volatile homophily.

%% file: related_work.tex
\section{Related Work}\label{sec:related-work}
\glspl{gcn} can broadly be categorized as either spatial or spectral.

Spectral \glspl{gcn}~\citep{bruna2014, chebnet, gcn, cayleynet} are based on graph signal processing theory, which makes them theoretically well-founded although only for undirected graphs.
This makes them ill-suited for road networks that often contain one-way streets and where, e.g., the driving speed on a road segment may heavily depend upon the direction of travel.
In addition, spectral \glspl{gcn} are transductive and rely on the specific graph structure on which it was trained and therefore does not support insertion and removal of edges to the network. This is problematic in road networks, where construction work frequently changes the network structure when, e.g., a motorway is built or a residential area is expanded.

Unlike spectral \glspl{gcn}, our proposed method explicitly considers road segment driving directions and is robust to substantial changes in the road network structure.
As we demonstrate in our experiments, our proposed method may make predictions on a different road network than it was trained on with only a minor decrease in predictive accuracy.

Our proposed method belongs to the class of spatial \glspl{gcn}~\citep{graphsage, gat}, which are inductive methods.
Like our proposed method, spatial \glspl{gcn} support directed graphs and changes in the network structure.
However, neither state-of-the-art spatial~\citep{graphsage, gat} nor spectral \glspl{gcn}~\citep{bruna2014, chebnet, gcn, cayleynet} support only node attributes.
In addition, both classes of \glspl{gcn} implicitly assume gradual changes in, e.g., driving speeds, when making predictions over the network and are thus ill-equipped to address the volatile homophily present in road networks.
In contrast, our method can incorporate edge attributes and between-edge attributes, in addition to node attributes, and has built-in mechanisms to address volatile homophily in the input road network.

Research in \glspl{gcn} architectures for road networks has thus far focused on extending \glspl{gcn} architectures to solve specific problems with temporal dependencies~\citep{spatio-temporal-gcn, stochastic-weight-completion,t-gcn}. In contrast, our work explores the spatial and structural aspects of \gls{gcn} architectures and we present a novel \gls{gcn} architecture that is designed to be generally applicable for machine learning on road networks. In addition, previous work has focused on spectral \gls{gcn} architectures, whereas our method is a spatial \gls{gcn} architecture. As discussed above, spatial \glspl{gcn} (such as our proposed method) have several advantages over spectral \glspl{gcn}, and we expect that our method can be used as a drop-in replacement for the spectral graph convolutions used in previous work.

Previous work has studied homophily in road networks~\citep{workshop}. Like us, they find that homophily is expressed differently in road networks---which we refer to as volatile homophily---but study this phenomenon in the context of transductive network embedding methods, whereas we focus on inductive \glspl{gcn}.
Unlike us, they do not propose a method to address the challenges of volatile homophily in road networks.

A preliminary four-page conference version of this paper presents the RFN-N+A variant~\citep{rfn-poster}.
The present version proposes an additional fusion function, $\textsc{InteractionalFuse}$, and an attentional aggregation function for RFNs.
These extensions yield substantial improvements in our experiments.
In addition, this version includes an evaluation of the \gls{rfn} variants’ robustness to changes in road network structure in cross-network inference, and it includes a detailed case study on the behavior of \glspl{rfn} graph convolutions under conditions of volatile homophily.

%% file: epilogue.tex
\section{Conclusion}\label{sec:conclusion}
We propose a method for machine learning on road networks.
We argue that many built-in assumptions of existing proposals do not apply in this setting, in particular the assumption of smooth homophily.
In addition, state-of-the-art \glspl{gcn} can leverage only one source of attribute information, whereas we identify three sources of attribute information in road networks: intersection, segment, and between-segment attributes.
We therefore propose the \emph{\glsfirst{rfn}}, a novel type of \gls{gcn} for road networks.

We compare different \gls{rfn} variants against a range of baselines, including two state-of-the-art \gls{gcn} algorithms on both within-network and cross-network driving speed estimation and speed limit classification.
We show that \glspl{rfn} are able to outperform the \gls{gcn} baselines by $32\text{--}40\%$ and $21\text{--}24\%$ on within-network driving speed estimation and speed limit classification, respectively.
Our experiments suggest that the assumptions of \glspl{gcn} do not apply to road networks: on within-network speed limit classification, the \gls{gcn} baselines fail to outperform a multi-layer perceptron that does not incorporate information from adjacent edges.
In addition, we show that \glspl{rfn} generalize better on cross-network driving speed estimation and suffered a $23\%$ smaller performance loss compared to the best performing neural network baseline.
We also show that \glspl{rfn} are more robust to volatile homophily than state-of-the-art \glspl{gcn} in a case study.

We use relatively small road networks, and we therefore find it prudent to comment on the scalability of \glspl{rfn}.
Both \glspl{rfn} and \glspl{gcn} have a worst time complexity of $O(|E|d^K)$, where $d = \max(\{|N(e) \mid e \in E\})$ is the max node degree of the dual graph and $K$ is the number of layers.

Several methods have been proposed to improve \gls{gcn} forward propagation efficiency.
First, by performing forward propagation only on the subnetwork necessary to compute an output for the road segments $E' \subseteq E$ for which an output is required~\citep{graphsage}.
Second, the global gradient can be approximated without recursive neighborhood expansion~\citep{fastgcn, asgcn, layer-dependent-sampling}.
To the best of our knowledge, \glspl{rfn} are compatible with these methods, and they can reduce the complexity to $O(|E'|dK)$ and $O(|E'|d^K)$ during training and inference, respectively.
In practice, $|E'|$ is often substantially smaller than $|E|$.
For instance, $|E'| = 5\,266$ and $|E| = 35\,947$ during testing for speed limit classification in Aalborg.
Hence, we expect that \glspl{rfn} can scale to country-sized road networks.

\subsection{Future Work}
Our experiments show that graph convolutional networks designed for road networks can yield substantial improvements in predictive performance.
\glspl{rfn} focus on the properties of volatile homophily, but do not explicitly utilize the spatial representation of road networks beyond road segment connectivity.
We expect that a more complete utilization of the spatial representations of road networks may yield further improvements over \glspl{gcn} for general networks.

\REVISED{
The \gls{rfn} variants we propose in this work cannot explicitly exploit temporal aspects of the data, but many tasks, e.g., driving speed estimation, are time-dependent. 
Extending \glspl{rfn} to learn temporal road network dynamics is therefore an important future direction.

Current spatio-temporal \gls{gcn} approaches focus on solving specific tasks by applying a temporal layer atop the spatial \gls{gcn} layer~\citep{spatio-temporal-gcn, stochastic-weight-completion,t-gcn}. As mentioned in \cref{sec:related-work}, we expect that the \gls{rfn} can be used as a drop-in replacement for the \gls{gcn} in such methods, but the \gls{rfn} architecture enables novel approaches for capturing temporal dynamics. For instance, the relationships between two adjacent road segments may change during the day. \glspl{rfn} may be able capture such changes, e.g., through time-dependent fusion functions that accept temporal inputs, thus making the behavior of the graph convolution itself time-dependent.
}

\COMMENT{
The \glspl{rfn} we propose in this work cannot explicitly exploit temporal aspects of the data.
Many tasks, e.g., driving speed estimation in our experiments, are time-dependent. 
Extending \glspl{rfn} to learn temporal road network dynamics, e.g., through time-dependent fusion functions that accept temporal inputs, is an important future direction.
}

All models achieved reasonable results on within-network speed limit classification, but failed to generalize to an unseen road network on cross-network speed limit classification.
However, all models also struggled to achieve the within-network results: the validation score could change drastically from epoch to epoch, causing us to employ early stopping as a regularization method in our experiments.
We therefore expect performance to be generally poor in areas that have not been observed during training. 
The speed limit data is crowd-sourced and thus labels tend to be concentrated in densely-populated areas leaving other areas scarcely labeled.

The driving speeds follow a similar pattern as the speed limits, where popular areas tend to have many driving speed observations, but we do not observe as severe a decline in predictive performance for cross-network driving speed estimation.
We hypothesize this is because a larger part of the road network is seen during training on this task, and because the decision boundaries in classification tasks make classification models more sensitive to changes in the input.
Investigating when knowledge learned from one part of a road network can be transferred to another part represents an interesting direction for future work.

\REVISED{
Finally, we have only investigated the use of \glspl{rfn} for machine learning on road networks.
However, \glspl{rfn} may be useful in other important application areas, e.g., multi-modal data fusion~\citep{multi-modal-attention,crowdstory} or machine learning on biological, citation, and social networks~\citep{graphsage,gat}. Investigating the applicability of \glspl{rfn} and their components to other types of networks and tasks present an interesting future direction.
}




%% file: acknowledgements.tex
\section*{Acknowledgments}
  The research presented in this paper is supported in part by the DiCyPS (Center for Data-Intensive Cyber-Physical Systems) project and by grants from the Obel Family Foundation and VILLUM FONDEN.\@
We also thank the \glsfirst{osm} contributors who have made this work possible.
Map data is copyrighted by the OpenStreetMap contributors and is available from {https://www.openstreetmap.org}.

%% file: appendix.tex
\section{Feature Derivation}\label{app:feature-derivation}
We describe here which node, edge, and between-edge features are used in our experiments and how they are derived from node, edge, and between-edge attributes.

\subsection{Node Attributes}
We derive node attributes using the zone map from the Danish Business Authority mentioned in \cref{sec:data-set}.
The map categorizes regions in Denmark as city zones, rural zones, and summer cottage zones.
We use a node attribute for each of the three zone categories to indicate whether or not an intersection belongs to the zone category or not.
Each of these node attributes is encoded as a binary value, either $0$ or $1$, yielding three features per node.

\subsection{Edge Attributes}
The \gls{osm} data includes categories for all road segments.
There are nine different road segment categories in the \gls{osm} extract that we use for our experiments.
In addition, the length of each road segment can be derived from its spatial representation.
We use both road segment category and length as edge attributes.

Road segment categories are one-hot encoded into nine-dimensional vectors.
Road segment lengths are encoded as continuous values that are scaled to the $[0; 1]$ range to ensure that all features are on the same scale.
Scaling road segment lengths into the same range as the other features makes training with the gradient-based optimization algorithm used in our experiments more stable.
The encoding of road segment categories and lengths yields a total of ten edge features.

Finally, the node features of the source and target nodes of the edge are concatenated with these edge features, yielding $16$ edge features in total.
The concatenation of source and target node features offers the neural network baselines (MLP, GraphSAGE, GAT) access to both node and edge features, rather than just edge features.

\subsection{Between-Edge Attributes}
The between-edge attributes we consider are the turn direction and the turn angle.

The turn direction can take on four values: straight-ahead, left, right, and U-turn.
We one-hot encode the turn-direction attribute into a four-dimensional vector.
The turn angle takes on values between $0$ and $180$ degrees.
We encode the turn angle attribute as a continuous value.
As with road segment lengths, turn angles are scaled into the $[0; 1]$ range.
This yields five between-edge features in total.

\section{Hyperparameter Selection}\label{app:hyperparam-selection}
As mentioned in \cref{sec:experimental-setup}, hyperparameter values for the neural network baselines are selected using a grid search over a set of hyperparameter configurations. 
For each algorithm, we select the hyperparameter configuration that achieves the best mean performance across ten runs.
We explore learning rates $\lambda \in \{0.1, 0.01, 0.001\}$ in the grid search based on preliminary experiments.
For GraphSAGE, GAT, and all \gls{rfn} variants, we explored $d \in \{32, 64, 128\}$ output dimensionalities of the first layer.
The MLP uses considerably fewer parameters than the other algorithms, and we therefore also explore larger hidden layer sizes $d \in \{128, 256, 512\}$ for a fair comparison.

\subsection{\gls{rfn} Variants}
For driving speed estimation, we use $L_2$ normalization on the first layer.
For speed limit classification, we found that training became more stable when using $L_2$ normalization on both layers.

\subsection{GraphSAGE}
GraphSAGE uses a pooling network to perform neighborhood aggregation.
For each layer in GraphSAGE, we set the output dimension of the pooling network to be double the output dimension of the layer in accordance with the authors' experiments~\citep{graphsage}.

By definition, GraphSAGE applies $L_2$ normalization on the output of each layer.
However, we omit $L_2$ normalization on the last (second) layer of GraphSAGE models for driving speed estimation; otherwise, the output cannot exceed a value of one, resulting in poor performance.

\subsection{GAT}
The GAT algorithm uses multiple attention heads that each output $d$ features at the first hidden layer.
These features are subsequently concatenated, thus yielding an output dimensionality of $h \cdot d$, where $h$ is the number of attention heads.
Based on past work~\citep{gat}, we explore different values of $h \in \{1, 2, 4, 8\}$ during the grid search and use the same number of attention heads for both layers.
The concatenation makes the GAT network very time-consuming to train when both $h$ and $d$ are large.
We therefore budget these parameters s.t.\ $h \cdot d \leq 256$, corresponding to, e.g., a GAT network with $64$ output units from $4$ attention heads.

\subsection{Selected Hyperparameters}
The hyperparameters selected used for evaluation on the test set is shown in \cref{tab:selected-hyperparams} for the Driving Speed Estimation (DSE) and Speed Limit Classification (SLC) tasks.

\begin{table}[ht]
  \centering
  \caption{\REVISED{Best model hyperparameters found in the grid search.\label{tab:selected-hyperparams}}}
  \addtolength{\tabcolsep}{-0.1cm}
  \scriptsize
  \REVISED{
  \begin{tabular}{lllllll}
    \toprule
    & \multicolumn{6}{c}{\emph{Task}} \\
    \cmidrule(lr){2-7}
    \emph{Algorithm} & \multicolumn{3}{c}{DSE} & \multicolumn{3}{c}{SLC} \\
    \midrule
    MLP & $d=128$ & $\lambda=0.01$  &       &
          $d=128$ & $\lambda=0.1$   &       \\
    GAT & $d=32$  & $\lambda=0.01$  & $h=8$ &
          $d=32$  & $\lambda=0.001$ & $h=8$ \\
    GraphSAGE & $d=64$ & $\lambda=0.01$  && 
                $d=64$ & $\lambda=0.001$ & \\
    \midrule
    RFN-N+A & $d=32$ & $\lambda=0.01$ &&
              $d=64$ & $\lambda=0.1$ & \\
    RFN-A+A & $d=32$ & $\lambda=0.01$ &&
              $d=32$ & $\lambda=0.1$  & \\
    RFN-N+I & $d=32$ & $\lambda=0.01$ &&
              $d=32$ & $\lambda=0.01$ & \\
    RFN-A+I & $d=32$ & $\lambda=0.01$ &&
              $d=64$ & $\lambda=0.01$ & \\
    \bottomrule
  \end{tabular}
}
  \addtolength{\tabcolsep}{-0.1cm}
\end{table}

\section{Case Study: Dall}\label{app:dall-case}
\begin{figure}[h!]
  \centering
  \includegraphics[width=\columnwidth]{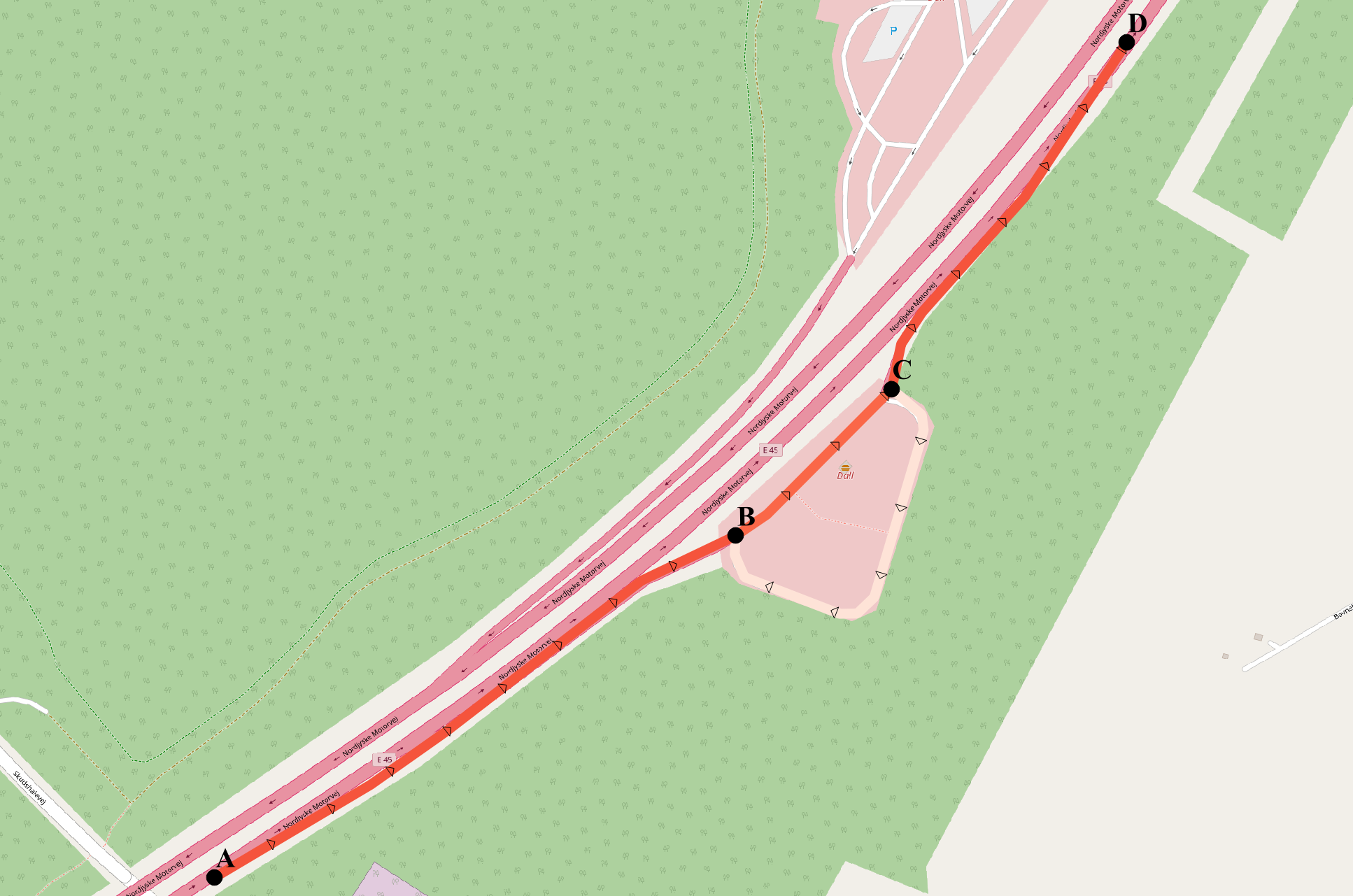}
  \caption{\REVISED{Ground truth driving speeds of a resting area and its adjacent motorway segments near the village of Dall, Denmark. Triangles indicate direction, black dots mark nodes, and color indicates speed: the darker, the faster.\label{fig:volatile-homophily-dall}}}
  \vspace{-0.25cm}
\end{figure}
The Danalien case discussed in \cref{sec:danalien-case} is a common case of volatile homophily, where there is still homophily s.t.\ each road segment has a similar neighbor segment.
However, as discussed in \cref{sec:introduction}, extreme cases of volatile homophily exists where there is no homophily.
Such a case is illustrated in \cref{fig:volatile-homophily-dall} that concerns a resting area next to a motorway near the village of Dall.
In the figure, segment CB is part of the resting area.
Segments AB and CD is a motorway approach and exit, respectively. 
Finally, segment BC is a transition segment that allows access to and from the resting area.
As indicated in \cref{fig:volatile-homophily-dall}, segment CB has a substantially different driving speed than the other segments.

\begin{figure}[ht]
  \vspace{-0.1cm}
  \centering
  \definecolor{transportation}{RGB}{255, 127, 14}
  \definecolor{transition}{RGB}{44, 160, 44}
  \definecolor{residential}{RGB}{23, 190, 207}
  \scriptsize
  \begin{tabular}{>{\centering\arraybackslash}m{0.67in} c >{\centering\arraybackslash}m{0.67in} c >{\centering\arraybackslash}m{0.67in} >{\centering\arraybackslash}m{0in}}
     \parbox{0.1\textwidth}{\centering\cellcolor{transportation} \textbf{Motorway Approach/Exit}} & &

    \parbox{0.1\textwidth}{\centering\cellcolor{transition} \textbf{Transition}} & &
    \parbox{0.1\textwidth}{\centering\cellcolor{residential} \textbf{Rest Area}} & \\[0.35cm]
    \vspace{-0.4cm}
  \end{tabular}
  \normalsize
  \begin{subfigure}{0.49\columnwidth}
    \includegraphics[width=\textwidth, trim={1.1cm, 0.75cm, 1.475cm, 1.225cm}, clip]{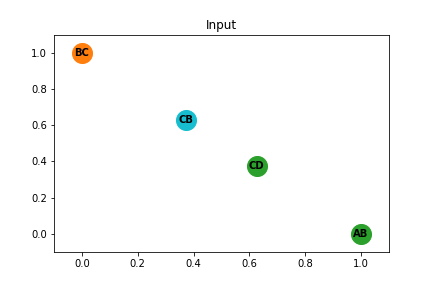}
    \caption{\REVISED{\scriptsize Input Representations}}
    \vspace{0.1cm}
  \end{subfigure}
  \begin{subfigure}{0.49\columnwidth}
    \includegraphics[width=\textwidth, trim={1.1cm, 0.75cm, 1.475cm, 1.225cm}, clip]{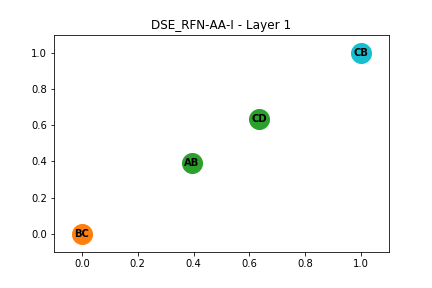}
    \caption{\REVISED{\scriptsize RFN-A+I}}
    \vspace{0.1cm}
  \end{subfigure}
  \begin{subfigure}{0.49\columnwidth}
    \includegraphics[width=\textwidth, trim={1.1cm, 0.75cm, 1.475cm, 1.225cm}, clip]{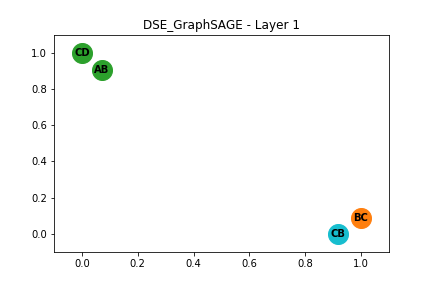}
    \caption{\REVISED{\scriptsize GraphSAGE}}
  \end{subfigure}
  \begin{subfigure}{0.49\columnwidth}
    \includegraphics[width=\textwidth, trim={1.1cm, 0.75cm, 1.475cm, 1.225cm}, clip]{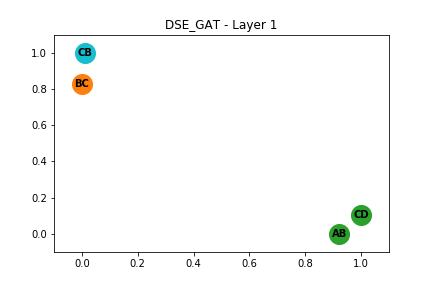}
    \caption{\REVISED{\scriptsize GAT}}
  \end{subfigure}
  \caption{\REVISED{\glsname{tsne} visualizations of the edge representations in the Dall case (a) before and (b, c, d) after the first graph convolution in different models.\label{fig:aggregations-dall}}}
  \vspace{-0.35cm}
\end{figure}

We select the \gls{rfn}, GraphSAGE, and GAT models and visualize the representations of the road segments of the Dall case before and after the first graph convolution in \cref{fig:aggregations-dall} following the same procedure as in the Danalien case (cf.\ \cref{sec:danalien-case}).

